\documentclass[runningheads]{llncs}

\usepackage{eccv}

\usepackage{eccvabbrv}

\usepackage{graphicx}
\usepackage{booktabs}
\usepackage[table]{xcolor}
\usepackage[accsupp]{axessibility}  

\usepackage{hyperref}

\usepackage[dvipsnames]{xcolor}  

\usepackage{orcidlink}

\begin{document}

\title{AgriField-40K: Adapting Vision Models to Agriculture With Efficient Continual Pretraining}

\titlerunning{AgriField-40K: Continual Pretraining for Agriculture}

\author{Vasileios Tzouras\inst{1,2}\orcidlink{0009-0002-7270-7032} \and Paraskevas Pegios\inst{1,2}\orcidlink{0009-0005-1471-4850}\and  
Lazaros Nalpantidis\inst{1,2}\orcidlink{0000-0002-3620-4123}}

\authorrunning{V.~Tzouras et al.}

\institute{Technical University of Denmark (DTU), Kongens Lyngby, Denmark\\
\email{\{vatzo,ppar,lanalpa\}@dtu.dk}
\and Pioneer Centre for Artificial Intelligence, Copenhagen, Denmark 
}

\maketitle

\begin{figure*}
    \centering
    \includegraphics[width=0.99\textwidth]{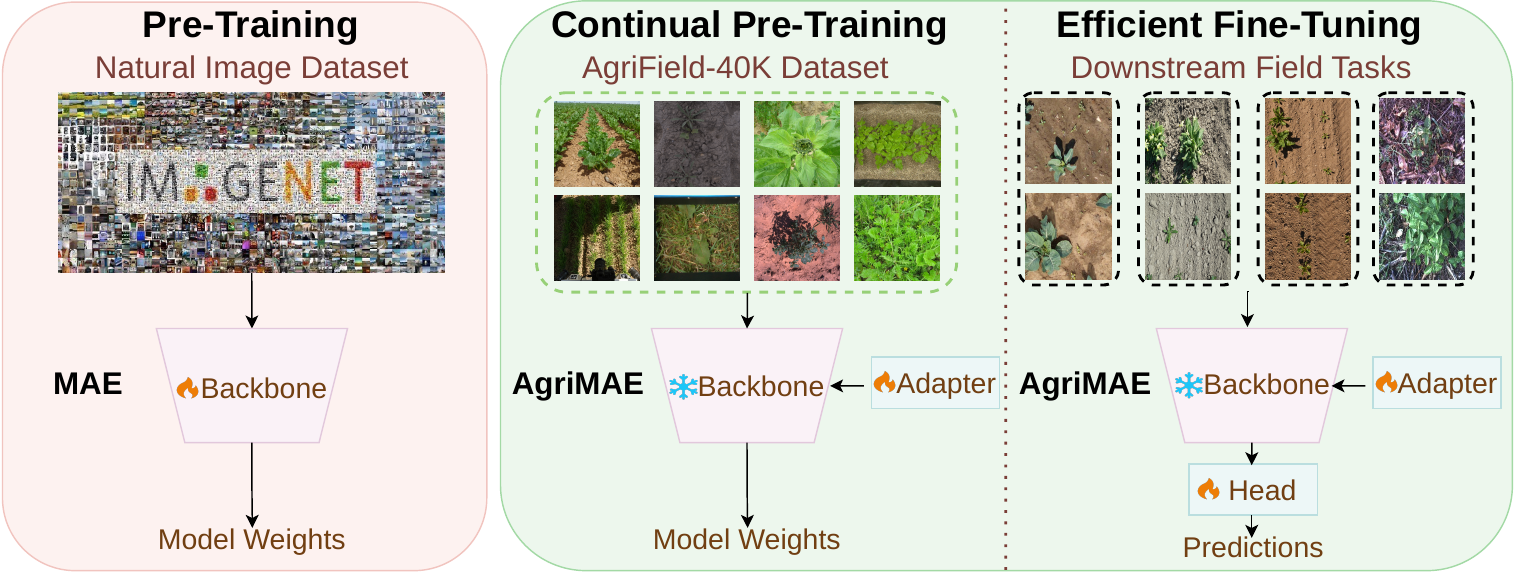}
    \caption{\textbf{Overview of our parameter-efficient adaptation framework.} Starting from MAE pretrained on natural images, AgriMAE performs continual pretraining on AgriField-40K by training only adapters. Then, we fine-tune these together with task-specific heads for downstream tasks, while keeping the backbone frozen.}
    \label{fig:teaser}
    \vspace{-25px}
\end{figure*}

\begin{abstract}

Field-based agricultural computer vision is important for precision agriculture, yet it largely depends on expensive annotations and costly adaptation of large pretrained models. We introduce AgriField-40K, a field-centric dataset curated from 17 public resources and covering diverse crops, weeds, and field conditions. Building on this, we present AgriMAE, a parameter-efficient continual pretraining baseline that adapts a masked autoencoder pretrained on natural images by training only lightweight adapters. We further explore semantic feature reconstruction as an alternative pretraining objective and evaluate transfer across multiple tasks. AgriMAE consistently improves downstream performance and can match or even outperform full fine-tuning while using up to $9\times$ fewer trainable parameters, showing that AgriField-40K is a practical resource for continual pretraining in agricultural vision. Project page: \url{https://dtu-pas.github.io/agrifield40k/}

\end{abstract}

\section{Introduction}
\label{sec:Introduction}
Computer vision is becoming an important tool for visual understanding in agricultural fields~\cite{tian2020computer}. It supports tasks such as crop and weed recognition~\cite{milioto2018real}, plant disease detection~\cite{mohanty2016using}, stress monitoring~\cite{singh2018deep}, fruit counting~\cite{bargoti2017deep}, phenotyping~\cite{rumexleaves}, and field monitoring~\cite{kamilaris2018deep}. These tasks can help reduce manual labour and support more precise and efficient farming. However, field-based agricultural vision still relies heavily on supervised learning, which requires large labelled datasets~\cite{data}. In practice, such labels are difficult to obtain as annotation requires domain expertise and must account for complex field conditions~\cite{guldenringfew}. 
At the same time, large amounts of field imagery are becoming easier to collect through drones, field robots, handheld cameras, and phenotyping systems~\cite{phenotyping}. This makes self-supervised learning (SSL) a promising direction for agricultural vision~\cite{guldenring2021self}.

SSL methods leverage strong visual priors learned from large-scale unlabelled data~\cite{simclr,moco,byol,simsiam,dino,barlow_twins,vicreg,swav}. Among these, masked image modelling (MIM)~\cite{mae,simmim,beit,ibot} has become a powerful paradigm for visual pretraining, with Masked Autoencoders (MAE)~\cite{mae} learning representations by reconstructing missing content from partially masked images. 
However, MAE is typically pretrained on natural-image datasets such as ImageNet~\cite{imagenet}, whereas agricultural field images differ substantially. The latter often contain repeated plant structures, fine-grained crop and weed differences, occlusion, changing illumination, and variation between growth stages, seasons, or acquisition platforms. This domain gap may reduce the effectiveness of features learned from natural-image pretraining when applied to downstream real-world agricultural field tasks.

A natural solution is to adapt existing foundation models through continual pretraining~\cite{agrifm}. Yet, fully adapting large models can be computationally expensive and may risk over-fitting to small datasets or degrading useful general-purpose representations~\cite{tzouras2025web}. Recent works~\cite{slr,explora,glare,pecop} have shown that continual pretraining with parameter-efficient fine-tuning (PEFT) mechanisms~\cite{houlsby,lora,uniadapter,adaptformer,vpt} can improve transfer to specialized domains under data or compute constraints. Despite this progress, \emph{parameter-efficient self-supervised continual pretraining} remains under-explored for field agricultural vision, largely due to the absence of a unified in-domain pretraining resource. Existing \emph{field data} is fragmented into many public datasets that are individually small in scale or limited in crop and weed coverage~\cite{deepweeds,rumexleaves,pnb,rn}, making it difficult to systematically develop and evaluate continual pretraining methods. 

To support the systematic study of efficient continual pretraining, we introduce \textbf{AgriField-40K}, a unified dataset constructed from 17 public field agricultural datasets and containing approximately 40,000 images. AgriField-40K brings together diverse real-field imagery, including crop-weed scenes, pasture vegetation, growth-stage variation, and broader field scenes. We further present \textbf{AgriMAE}, a strong parameter-efficient baseline for this setting. As illustrated in Fig.~\ref{fig:teaser}, AgriMAE adapts an ImageNet-pretrained MAE to AgriField-40K with lightweight adapters and transfers it to downstream tasks through PEFT. To analyse the effect of the pretraining objective in MIM, we compare standard pixel-level reconstruction~\cite{mae} with reconstruction of DINOv3~\cite{dinov3} representations, providing a practical reference point for agricultural vision. Finally, we demonstrate the effectiveness of our pipeline on downstream applications, including classification, semantic segmentation and object detection tasks.

Our main contributions are the following:
\begin{itemize}
    \item The first \emph{field-centric} agricultural dataset for representation learning and continual pretraining, \textbf{AgriField-40K}, built from 17 public datasets spanning diverse and field conditions and acquisition settings.
    
    \item A simple yet effective parameter-efficient baseline, \textbf{AgriMAE}, which adapts a standard MAE to agricultural field imagery by freezing the pretrained backbone and training lightweight adapters on our dataset.

    \item A study of reconstruction objectives for agricultural continual pretraining, comparing pixel-level reconstruction with semantic feature reconstruction.

    \item Downstream validation showing that our continual pretraining strategy consistently improves downstream performance in parameter-efficient settings.
\end{itemize}

\section{Related Work}
\label{sec:Related Work}

\noindent \textbf{Self-Supervised Learning.} Early SSL methods relied on contrastive learning~\cite{moco,simclr}, while later approaches avoided explicit negative pairs through redundancy reduction~\cite{barlow_twins,vicreg}, clustering~\cite{swav}, and self-distillation~\cite{byol,simsiam,dino}. Modern self-distillation models such as DINOv2~\cite{dinov2} and DINOv3~\cite{dinov3} learn highly transferable semantic features, but their success comes with substantial training cost, relying on large curated datasets, heavy architectures, and carefully scaled optimization pipelines. Another direction is MIM, inspired by natural language processing~\cite{bert}. BEiT~\cite{beit} predicts discrete visual tokens, while MAE~\cite{mae} simplifies masked prediction by reconstructing pixels with an asymmetric encoder-decoder architecture. Although simple and effective, pixel targets provide limited semantic supervision. To address this, I-JEPA~\cite{ijepa} predicts masked regions in latent space, iBOT~\cite{ibot} combines MIM with self-distillation using DINO~\cite{dino}, and MILAN~\cite{milan} reconstructs text-aligned CLIP~\cite{clip} image features which are often object-centric. In this work, we explore DINOv3~\cite{dinov3} as a feature extractor for semantic feature reconstruction, using its dense representations to provide feature-level guidance for field imagery where crop, weed, and background structure is often distributed across the scene. \\

\noindent \textbf{Efficient Continual Pretraining.} Pretrained vision models can be adapted to new domains by continuing SSL on unlabelled target-domain data~\cite{hpt,gfm,agrifm}, avoiding training from scratch that typically requires large datasets. Early methods continue training all backbone parameters, using either MAE~\cite{mae}-based masked modelling~\cite{gfm} or hierarchical SSL pretraining~\cite{hpt,agrifm} from an existing pretrained initialization. However, updating the full backbone can still be expensive for large models and may be unnecessary when the goal is domain adaptation. To further reduce this cost, recent works~\cite{slr,explora,glare,pecop} apply PEFT to continual pretraining. PEFT methods were originally introduced for supervised adaptation, where most backbone parameters are frozen and only a small set of modules are trained, such as bottleneck adapters~\cite{houlsby}, LoRA~\cite{lora}, visual prompts~\cite{vpt}, AdaptFormer~\cite{adaptformer}, or UniAdapter~\cite{uniadapter}. In the context of continual pretraining, parameter-efficient adaptation has been applied in several settings, with SLR~\cite{slr} for remote sensing, ExPLoRA~\cite{explora} for satellite, medical, and wildlife domains, GLARE~\cite{glare} for dense prediction, and PECoP~\cite{pecop} for video action quality assessment. Our work follows this direction, with AgriField-40K enabling continual pretraining research and applications in agriculture and AgriMAE providing a strong parameter-efficient baseline. \\

\noindent \textbf{Representation Learning in Agriculture.} Agricultural vision has adopted SSL methods including contrastive~\cite{sornapudi2024self,bunyang2023self} and prototype-based learning~\cite{guldenring2021self}, MIM~\cite{shikhar2024label,wang2024classification}, self-distillation~\cite{zhang2022self}, hybrid approaches~\cite{nagasubramanian2022plant,wang2024self}, and CLIP-style~\cite{clip} multimodal image-text alignment~\cite{cao2023cucumber,nawaz2025agriclip}. Yet, most studies pretrain on a single dataset or target a specific downstream task, such as disease classification~\cite{disease_classification} or fruit ripeness estimation~\cite{fruit_ripeness}. Recent work has also explored foundation models beyond task-specific pretraining. SPROUT~\cite{xiang2026sprout} explores diffusion-based representation learning, WeedNet~\cite{shen2025weednet} pretrains MAE on large-scale web images before adapting it to weed recognition, and Agri-FM+~\cite{agrifm} performs continual SSL pretraining from ImageNet-pretrained ResNet-50~\cite{resnet} on a large-scale agricultural corpus to improve downstream dense performance. Closer to our goal, Espejo \etal~\cite{espejo2025foundation} evaluate DINOv2 features with linear probing, full and efficient fine-tuning  across disease, weed, and growth-stage tasks, while Chen \etal~\cite{chen2023adapting} study adaptation of pretrained MAE~\cite{mae}, DINO~\cite{dino}, and DINOv2~\cite{dinov2} backbones for plant phenotyping using task-specific LoRA~\cite{lora} adapters.  However, these approaches adapt and evaluate models during downstream supervised training. In contrast, we present AgriMAE, which uses adapters for parameter-efficient continual pretraining on unlabelled images from AgriField-40K, to produce reusable representations for downstream real-world field tasks. \\

\noindent \textbf{Datasets for Agricultural Vision.}
Existing datasets cover a wide range of tasks, with many resources focusing on leaf-level, whole-plant, or close-up imagery captured under controlled conditions. PlantVillage~\cite{plantvillage} and PlantDoc~\cite{plantdoc} support leaf disease recognition, while CVPPP~\cite{cvppp} provides controlled leaf-level segmentation. Larger species-recognition resources such as Pl@ntNet-300K~\cite{plantnet}, iNaturalist~\cite{inatspecies}, PlantCLEF~\cite{plantclef}, and iNatAg~\cite{inatag} broaden taxonomic coverage, but are mainly oriented toward isolated plants or close-up species identification rather than canopy-level field monitoring. Field datasets better capture real-world visual conditions, including crop rows, soil clutter, mixed vegetation, and changing illumination, yet they are often limited to specific crops, weeds, platforms, or regions, such as sugar beet~\cite{pnb,sugarbeet1,sugarbeet2}, selected crop types~\cite{mustc}, single-crop or weed-focused datasets~\cite{rn,vcd,pags,rumexweeds,rumexleaves}, and rangeland weed imagery~\cite{deepweeds}. Closest to our work, Agri-147K~\cite{agrifm} combines multiple datasets for general agricultural pretraining. However, many of its sources remain leaf-centric, and, to the best of our knowledge, the curated corpus has not been publicly released. In this paper, we prioritize field imagery and carefully curate AgriField-40K from public datasets with permissive licenses, enabling release as a unified resource for parameter-efficient continual self-supervised pretraining.

\section{AgriField-40K Dataset}
\label{sec:agrifield}

\begin{figure*}[t]
    \centering
    \includegraphics[width=1.0\textwidth]{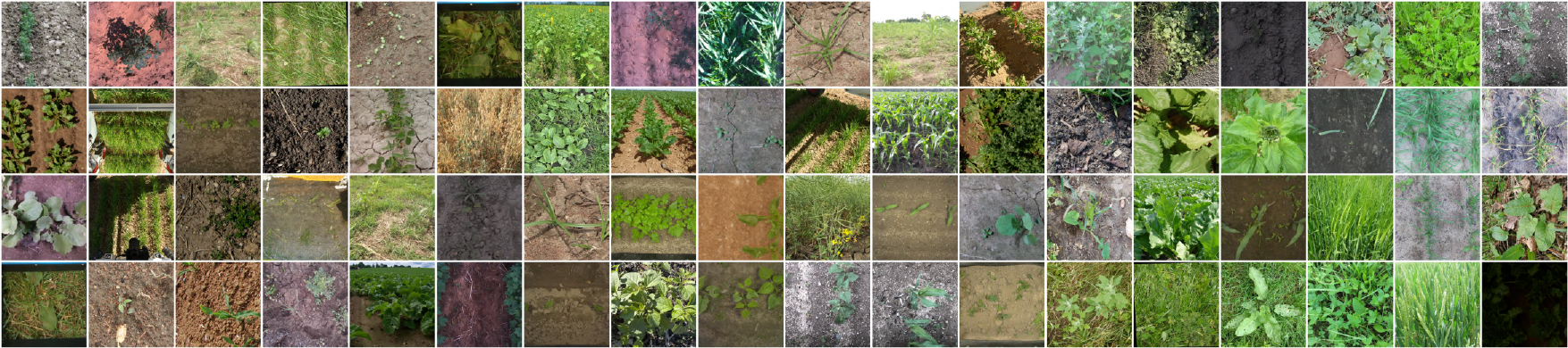}
    \caption{Sample images from our field-centric AgriField-40K dataset.}
    \label{fig:dataset}
\end{figure*}

We introduce \textbf{AgriField-40K}, a well-curated field-centric agricultural dataset for representation learning and continual pretraining built from 17 publicly available resources. As shown in Fig.~\ref{fig:dataset}, we define \emph{field-centric} imagery as agricultural imagery captured under \emph{real field conditions}. This includes crop and weed mixtures, pasture vegetation, dense canopies, soil backgrounds, different growth stages, and data collected from handheld cameras, robots, drones, and shrouded platforms. Table~\ref{tab:datasets} summarizes the resources, licenses, original and retained sizes, field domains, acquisition settings, and original tasks. \textbf{AgriField-40K} combines crop-focused resources such as MuST-C~\cite{mustc}, VegAnn~\cite{vgn}, LUCASVision~\cite{lvb}, and VCD~\cite{vcd}, which provide images across crop species, growth stages, camera setups, and geographic locations. It also includes crop-weed datasets such as PhenoBench~\cite{pnb}, WE3DS~\cite{weds}, ACRECropWeed~\cite{acw}, Ronin~\cite{rn}, SorghumWeed~\cite{swd}, and Maize-Weed~\cite{mw}, adding mixed vegetation, robot and patch-based drone views, and temporal or illumination variation. To increase fine-grained weed and dense-vegetation diversity, we further include RadishWheat~\cite{pags}, PalmerAmaranth~\cite{rwd}, RumexLeaves~\cite{rumexleaves}, Sesame\&Weed~\cite{saw}, PerennialPlants~\cite{per}, and GrassClover~\cite{gc}. Finally, we manually curate iNatWeeds, a set of field-like weed images collected from iNaturalist~\cite{inatweeds} under permissive licenses. To ensure license compatibility, \textbf{AgriField-40K} is released under CC BY-SA 4.0, following the most restrictive license. \\

\begin{table*}[t]
\centering
\caption{AgriField-40K Dataset Summary: Unless noted otherwise, CC BY denotes CC BY 4.0; VegAnn~\cite{vgn} uses CC BY 1.0. In the acquisition column, ``H. Cameras'' denotes handheld cameras and ``O. Cameras'' denotes overhead cameras.}
\label{tab:datasets}
\resizebox{\textwidth}{!}{%
\begin{tabular}{ccccccccc}
\hline
\textbf{Dataset} &
\textbf{Year} &
\textbf{License} &
\textbf{Size} &
\textbf{Retained} &
\textbf{Domain} &
\textbf{Acquisition} &
\textbf{Task} \\
\hline
\hline

&  &  &  &  
& Sugar Beet, Soybean, 
&  & \\

MuST-C~\cite{mustc} & 2026 & CC BY & 7,242 & 7,242
& Potato, Maize,
&  Robot &  -- \\

&  &  &  &  
& Wheat \& Intercrop
&  &  \\

\hline

VCD~\cite{vcd} & 2022 & CC BY & 2,258 & 2,258 & Maize, Bean \& Leek & Shrouded  & Detection \\
 &  &  &  &  & (Early Stage) & Platform  &   \\

\hline

PalmerAmaranth~\cite{pags} & 2023 & CC BY & 614 & 516 & Palmer Amaranth & H. Cameras & Detection \\
 &  &  &  &  & (8 Stages) &  &   \\

\hline

ACRECropWeed~\cite{acw} & 2023 & CC BY & 1,000 & 791 & Maize, Beans & Robot & Multi-Task \\
 &  &  &  &  & \& 4 Weeds &  &   \\

 \hline

 SorghumWeed~\cite{swd} & 2023 & CC BY & 252 & 172 & Sorghum, Grasses & H. Cameras & Multi-Task \\
 &  &  &  &  & \& Weeds &  &   \\
 
\hline

GrassClover~\cite{gc} & 2019 & CC BY-SA & 435 & 435 
& Grass, Clover  & H. Cameras & Segmentation \\

&  & &  &  & \& Weeds & \& Synthetic & \\
\hline

PhenoBench~\cite{pnb} & 2026 & CC BY-SA & 29,312 & 
9,606 & Sugar Beet \& 6 Weeds & Drone & Segmentation \\

\hline

VegAnn~\cite{vgn} & 2022 & CC BY & 3,775 & 1,607 & 26+ Crops & Multiple & Segmentation \\

\hline

Ronin~\cite{rn} & 2021 & CC BY  & 1,176 & 135 & 6 Crops \& 8 Weeds &  H. Cameras & Detection \\

\hline

LUCASVision~\cite{lvb} & 2023 & CC BY & 15,876 & 11,195 & 12 Crops & H.  Cameras & Classification \\

\hline

WE3DS~\cite{weds} & 2023 & CC BY & 2,568 & 1,553 & 7 Crops \& 10 weeds & Stereo RGB-D & Segmentation \\

\hline

Maize-Weed~\cite{mw} & 2022 & CC BY & 843 & 255 & Maize \&  Weeds & H. Cameras & Detection \\

\hline

RadishWheat~\cite{rwd} & 2022 & CC BY & 552 & 534 & Wild Radish in Wheat & O. Cameras & Detection \\

\hline

RumexLeaves~\cite{rumexleaves} & 2024 & CC BY & 809 & 809 & Rumex Obtusifolius & Robot & Detection \\

\hline

Sesame\&Weed~\cite{saw} & 2020 & CC0 & 1,300 & 1,300 & Sesame \& Weeds & H. Cameras & Detection \\

\hline

PerennialPlants~\cite{per} & 2021 & MIT & 392 & 240 & Weeds in Perennials & H. Cameras & Mult-Task \\

\hline

iNatWeeds & 2026 & CC BY & 1,315 & 1,315 & Mixed Species & H. Cameras & -- \\

\hline
\hline

\textbf{AgriField-40K} & \textbf{2026} & \textbf{CC BY-SA} & --- & \textbf{39,963} & \textbf{Field-Centric} & \textbf{Multiple} & \textbf{Pretraining} \\

\hline

\end{tabular}%
}
\end{table*}

\noindent\textbf{Data Preprocessing \& Filtering.}
Our goal is efficient continual pretraining. Therefore, we discard original annotations, such as segmentation masks or class labels, and retain only RGB images. We also rename images with a dataset-specific prefix such that sources remain recoverable. Several source datasets, especially those collected from video streams~\cite{mw} or robot platforms~\cite{rn}, contain consecutive frames that are visually near-identical. These could potentially bias the corpus and reduce its effective diversity. Thus, we apply fixed-interval sampling and retain one frame every $k$ frames, where $k$ is chosen based on the frame rate and observed visual change between consecutive frames. For all resources, we manually remove, blurry and low-resolution samples, and other low-quality images that provide limited useful signal for pretraining. \\

\noindent\textbf{Image Resizing.}
To standardize the input resolution while avoiding geometric distortion, each image is resized with its aspect ratio preserved and then centre-cropped. Given an image with width $W$ and height $H$, we rescale it so that the shorter side matches $T=512$ pixels using the scale factor $s=\frac{T}{\min(W,H)}$ and Lanczos interpolation~\cite{lanczos}. The resized image is then centre-cropped to a final resolution of $T \times T$. This produces a common input size while preserving the original image geometry, although small border regions may be removed. \\

\noindent\textbf{Train \& Validation Splits.}
We split AgriField-40K into 80\% training and 20\% validation images to monitor reconstruction quality during continual pretraining. For source datasets derived from image sequences, we split at the sequence level rather than the image level to prevent near-duplicate frames from the same sequence from appearing in both splits. For source datasets composed of independently captured images, we apply standard random splits. \\

\section{AgriMAE: Efficient Model Adaptation for Agriculture}

\begin{figure*}[t]
    \centering
    \includegraphics[width=1.0\textwidth]{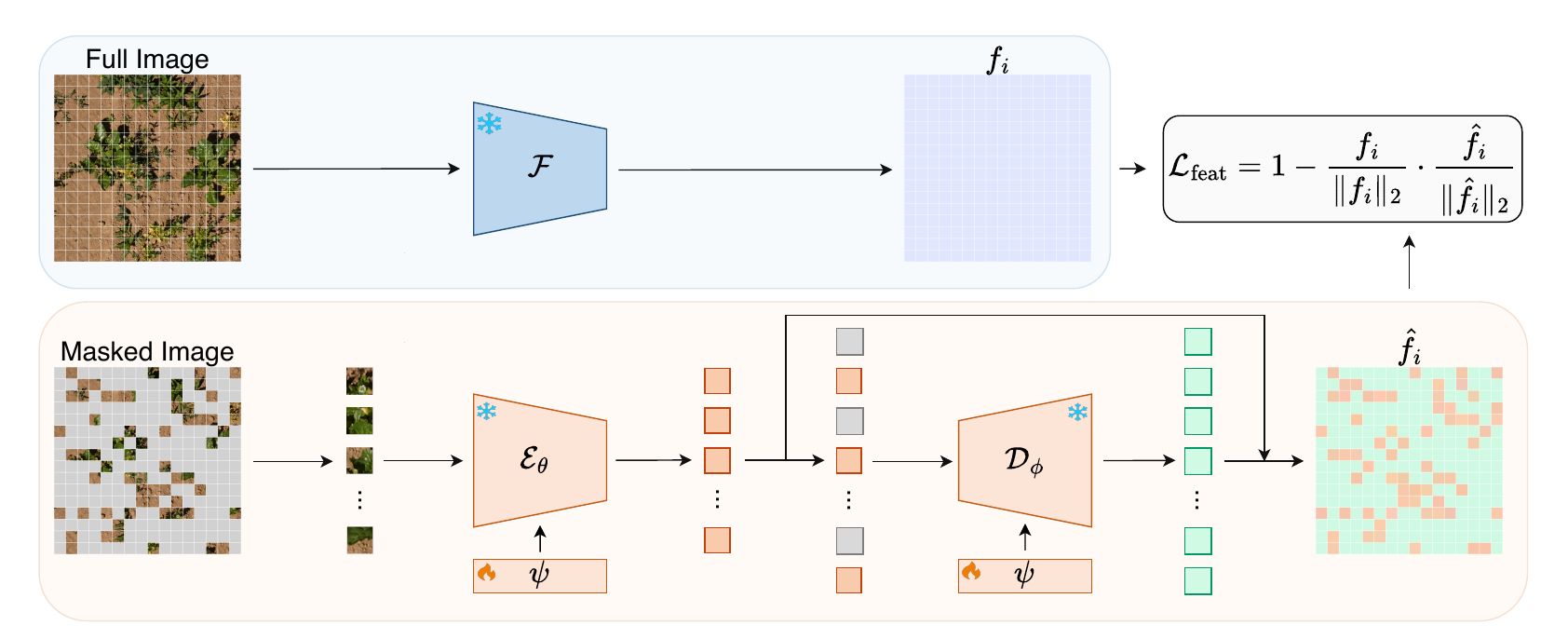}
    \caption{Overview of AgriMAE with semantic feature reconstruction. An ImageNet-pretrained MAE encoder-decoder (bottom), is adapted with lightweight AdaptFormer modules ($\psi$) on AgriField-40K to predict patch-level features from a frozen feature extractor $\mathcal{F}$ (top), with the loss computed over both masked and visible patches.}
    \label{fig:model}
\end{figure*}

We build on MIM with a vision transformer (ViT)~\cite{vit} backbone, following MAE~\cite{mae}, to establish a strong parameter-efficient baseline for continual pretraining. Our model adaptation pipeline from continual pretraining to downstream adaptation is shown in Fig.~\ref{fig:teaser}. We refer to our approach as \textbf{AgriMAE} and denote the encoder and decoder of the pretrained masked autoencoder by $\mathcal{E}_{\theta}$ and $\mathcal{D}_{\phi}$, respectively, where parameters $\theta$ and $\phi$ remain frozen. The trainable adapter parameters inserted into the MAE are denoted by $\psi$. In addition to the standard pixel-reconstruction objective, we explore semantic feature reconstruction using a strong frozen feature extractor, denoted by $\mathcal{F}$. During continual pretraining, only $\psi$ is updated using our \textbf{AgriField-40K} dataset, as illustrated in Fig.~\ref{fig:model}. During downstream adaptation, we employ the same principal through PEFT and further optimize $\psi$ together with a task-specific prediction head using labelled data for each task. 

\subsection{Parameter-Efficient Adaptation}
\label{sec:adaptformer}
For both continual pretraining on AgriField-40K and downstream PEFT, we use AdaptFormer~\cite{adaptformer}. More specifically, we insert a lightweight bottleneck adapter in parallel with the feed-forward multilayer perceptron (MLP) module of each transformer block, while keeping the original transformer weights frozen. Given a block input $\hat{X} \in \mathbb{R}^{n \times d}$, where $n$ is the number of tokens and $d$ is the feature dimension, the adapted block output is computed as
\begin{equation}
    Y = \hat{X} + \mathrm{MLP}(\mathrm{LN}(\hat{X})) 
    + s \cdot W_\mathrm{up}\,\sigma\!\left(W_\mathrm{down}\,\mathrm{LN}(\hat{X})\right),
    \label{eq:adaptformer}
\end{equation}
where $\mathrm{LN}(\cdot)$ denotes layer normalization, $\mathrm{MLP}(\cdot)$ is the frozen feed-forward module of the transformer block, $W_\mathrm{down} \in \mathbb{R}^{r \times d}$ and $W_\mathrm{up} \in \mathbb{R}^{d \times r}$ project features into and out of a bottleneck of rank $r \ll d$, $\sigma(\cdot)$ is a non-linear activation, and $s$ is a learnable scale. We denote the adapter parameters as $\psi=\{W_\mathrm{down}, W_\mathrm{up}, s\}$. During both stages, only $\psi$ and the task-specific heads used for downstream transfer are updated, while the pretrained backbone remains frozen.

\begin{figure}[t]
    \centering
    \begin{subfigure}{0.32\textwidth}
        \centering
        \includegraphics[width=\linewidth]{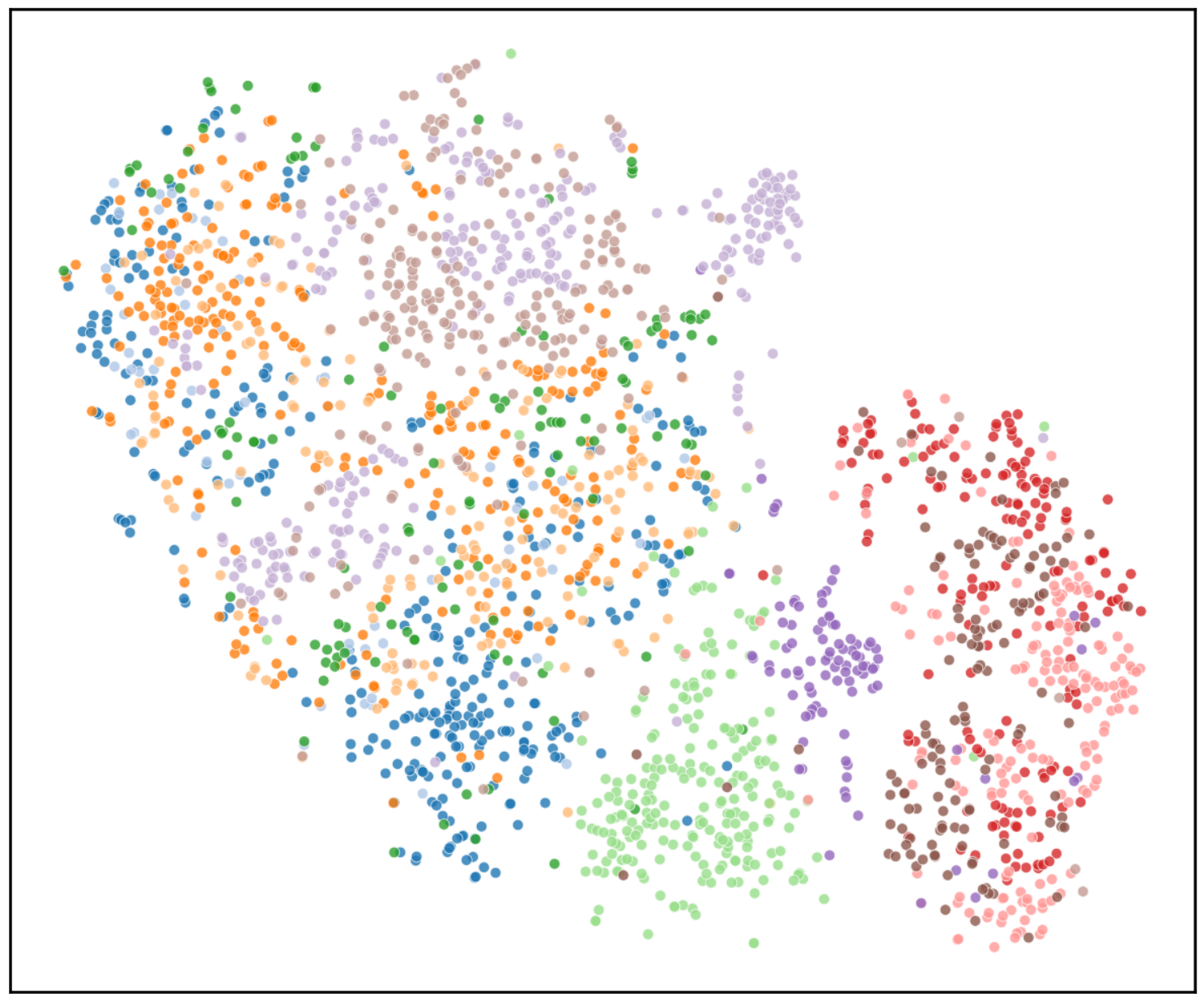}
        \caption{\tiny MAE}
        \label{fig:mae_pretrained}
    \end{subfigure}
    \hfill
    \begin{subfigure}{0.32\textwidth}
        \centering
        \includegraphics[width=\linewidth]{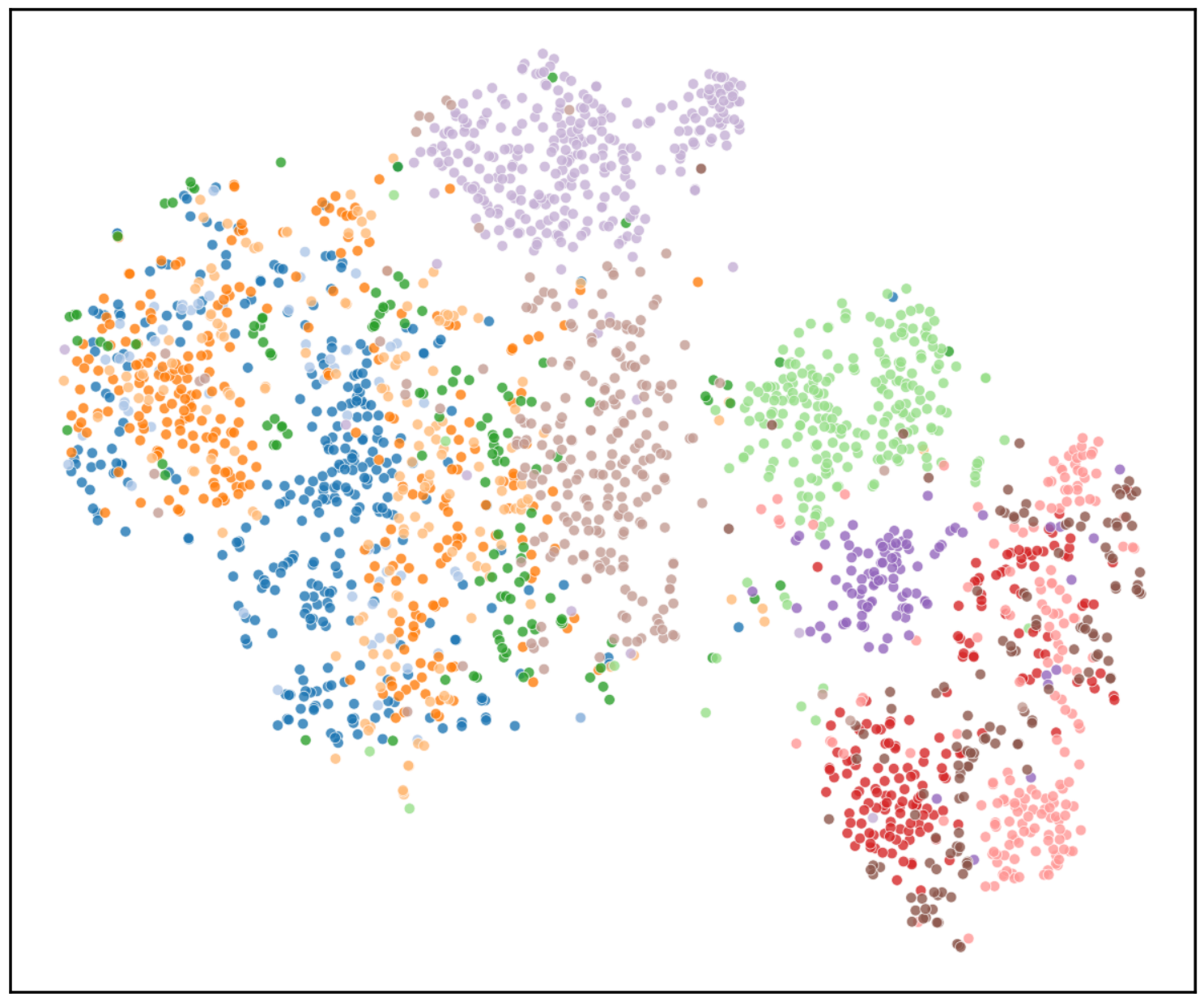}
        \caption{\tiny AgriMAE with $\mathcal{L}_{\mathrm{pix}}$}
        \label{fig:agrimae_pix}
    \end{subfigure}
    \hfill
    \begin{subfigure}{0.32\textwidth}
        \centering
        \includegraphics[width=\linewidth]{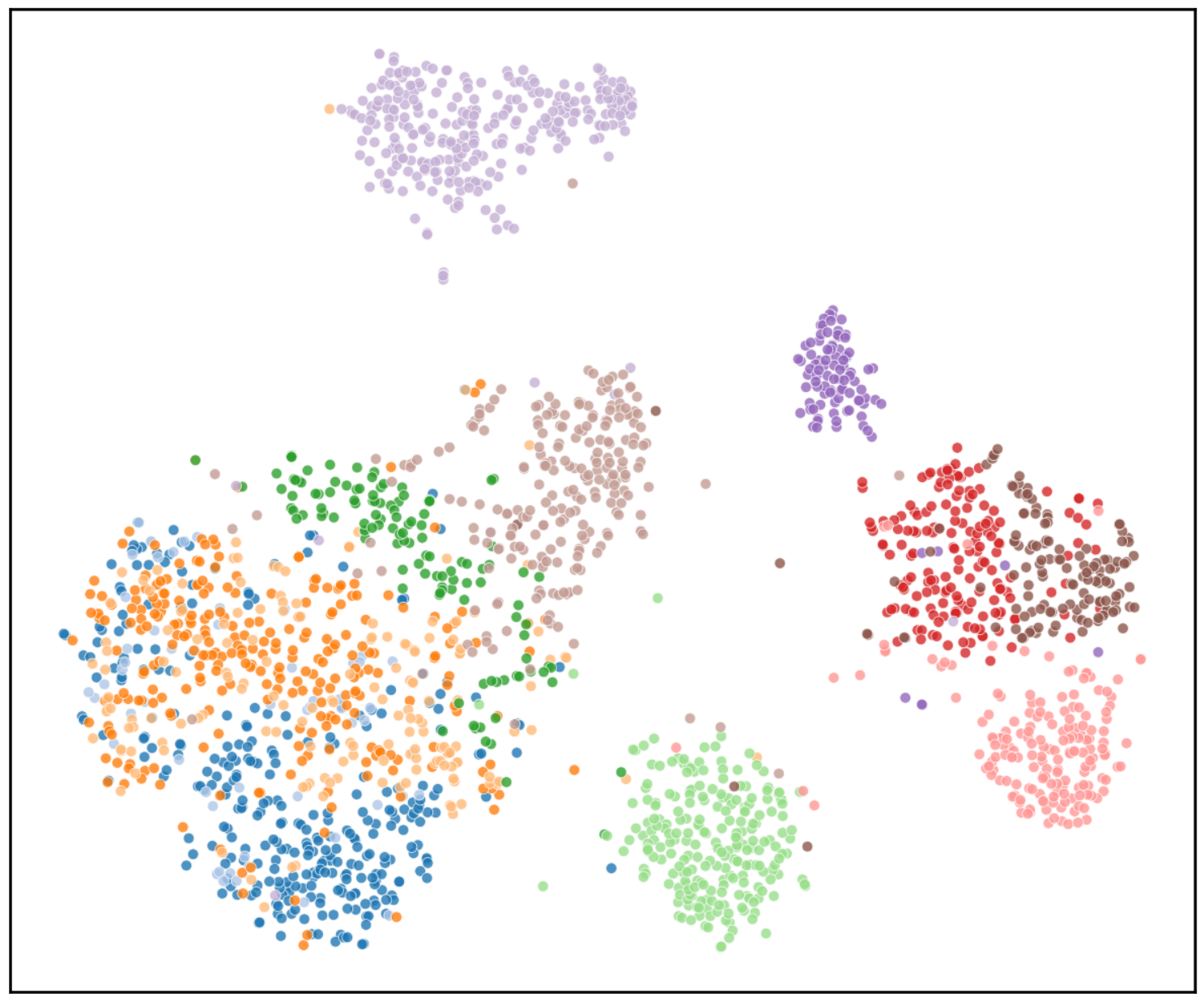}
        \caption{\tiny AgriMAE with $\mathcal{L}_{\mathrm{feat}}$}
        \label{fig:agrimae_feat}
    \end{subfigure}
    \caption{t-SNE~\cite{tsne} visualization of feature representations from different pretraining methods. Features are extracted from the final layer using images from the LUCASVision~\cite{lvb} subset of the AgriField-40K validation split.}
    \label{fig:tsne_comparison}
\end{figure}

\subsection{Efficient Self-Supervised Continual Pretraining}
Given an RGB image $x$, we divide it into $n$ non-overlapping patches $\{x_i\}_{i=1}^{n}$, each of size $P$. Each patch is linearly projected to dimension $d$ and combined with positional embeddings. A random binary mask $m \in \{0,1\}^n$ removes a fraction $\delta$ of the patches, and the encoder $\mathcal{E}_{\theta}$ processes only the visible tokens. The decoder $\mathcal{D}_{\phi}$ reconstructs the masked patches from the encoded visible tokens and learnable, position-aware \texttt{[MASK]} tokens. Denoting the decoder output by $\hat{x}$, the standard MAE objective minimizes the mean squared error between the reconstructed and ground-truth RGB pixel values at the masked positions:
\begin{equation}
    \mathcal{L}_{\mathrm{pix}} = \frac{1}{\delta \cdot n} \sum^{n}_{i=1} m_i \cdot | x_i - \hat{x}_i |^2.
    \label{eq:l_rec}
\end{equation}

We freeze the encoder $\mathcal{E}_{\theta}$ and decoder $\mathcal{D}_{\phi}$, and place an adapter, as defined in Eq.~\ref{eq:adaptformer}, in every transformer block. During continual pretraining, only adapter parameters $\psi$ are updated to minimize $\mathcal{L}_{\mathrm{pix}}$, keeping parameters $\theta$ and $\phi$ fixed.

\subsection{Semantic Feature Reconstruction}
\label{sec:feature_reconstruction}

We explore a semantic feature reconstruction objective using dense embeddings from a strong, frozen feature extractor $\mathcal{F}$. Given the full unmasked image $x$, $\mathcal{F}$ produces one target embedding for each patch, \ie, $\{f_i\}_{i=1}^{n}$. 
In parallel, the masked autoencoder receives only the visible patches under the mask $m$ and predicts a feature embedding for each patch. To predict features instead of pixel RGB values, we replace the decoder's pixel-prediction head with a linear projection into the feature space of $\mathcal{F}$. The decoder $\mathcal{D}_\phi$ outputs a sequence of predicted patch embeddings $\{\hat{f}_i\}_{i=1}^{n}$. 
We compare predicted and target embeddings using cosine distance after $\ell_2$ normalization:
\begin{equation}
    \mathcal{L}_{\mathrm{feat}} =
    \frac{1}{n}\sum_{i=1}^{n}
    \left(
    1 -
    \frac{f_i}{\lVert f_i \rVert_2}
    \cdot
    \frac{\hat{f}_i}{\lVert \hat{f}_i \rVert_2}
    \right).
    \label{eq:l_feat}
\end{equation}

Unlike $\mathcal{L}_{\mathrm{pix}}$, which is computed only at masked positions, we compute $\mathcal{L}_{\mathrm{feat}}$ over all patches at both masked and unmasked positions. This encourages AgriMAE to reconstruct semantic structure across the whole image rather than only missing low-level pixel content. Thus, AgriMAE can be trained with either pixel reconstruction loss in Eq.~\ref{eq:l_rec} or semantic feature reconstruction loss in Eq.~\ref{eq:l_feat} as the continual pretraining objective.

Figure~\ref{fig:tsne_comparison} visualizes t-SNE~\cite{tsne} embeddings from the LUCASVision~\cite{lvb} subset of the AgriField-40K validation set, comparing the original ImageNet-pretrained MAE with AgriMAE trained using pixel and feature reconstruction. The standard MAE produces highly mixed clusters, reflecting the domain gap between natural images and agricultural field imagery. Continual pretraining with pixel reconstruction already improves the embedding structure, while feature reconstruction leads to more compact and separable clusters, preserving meaningful similarities among related crop categories such as wheat, durum wheat, and rye.

\section{Experiments and Results}
\label{sec:Experiments}

We implement \textbf{AgriMAE}, using a ViT-B/16~\cite{vit} model pretrained with standard MAE~\cite{mae} pixel reconstruction objective on ImageNet-1K~\cite{imagenet}. During continual self-supervised pretraining on \textbf{AgriField-40K}, only adapter parameters are optimized, while the encoder and decoder remain frozen. We use a masking ratio of 0.75 and extract feature reconstruction targets from a frozen DINOv3 ViT-L/16 model pretrained on LVD-1689M~\cite{dinov3}. Optimization uses AdamW~\cite{adamw} with a learning rate of $1\times10^{-4}$, weight decay of $1\times10^{-2}$, and $\beta=(0.9,0.95)$, with a 5-epoch linear warm-up followed by cosine decay for a total of 200 epochs and an effective batch size of 128. Unless stated otherwise, we use adapter bottleneck rank $r=512$ and standard MAE augmentations, \eg, random resized crop and horizontal flip. All continual pretraining experiments are conducted on a single consumer-grade NVIDIA RTX 5090 GPU.  
\\

\subsection{Downstream Tasks and Datasets}
To evaluate the effectiveness of our pipeline for \emph{field-centric} agricultural vision, we benchmark it across diverse downstream tasks using four real-world datasets. We consider classification, semantic segmentation, and object detection benchmarks, covering weed recognition, crop-weed segmentation, plant-level field segmentation, and multi-species crop-weed detection. Our evaluation focuses on \emph{efficient fine-tuning}, where the backbone remains frozen and only adapters and task-specific prediction heads are optimized, allowing us to isolate the effect of continual pretraining in a parameter-efficient setting. We compare: (i) randomly initialized adapters inserted into the frozen MAE encoder, (ii) AgriMAE adapters continually pretrained with either pixel or feature reconstruction, and (iii) full fine-tuning of the original MAE encoder as a strong reference point. \\

\noindent \textbf{Classification.}
We use DeepWeeds~\cite{deepweeds}, which consists of 17,509 RGB images from nine classes: eight weed species in Australian rangelands and one background class. To reduce class imbalance, we subsample the background class, which accounts for over 50\% of the samples. The data is split into 60\%/20\%/20\% train, validation, and test sets. We evaluate both adapter fine-tuning and linear probing. For adapter fine-tuning, we freeze the pretrained backbone and fine-tune the adapters and a linear classification head using AdamW with learning rates of $1\times10^{-5}$ and $1\times10^{-4}$, respectively, a weight decay of $1\times10^{-2}$, for 100 epochs with a batch size of 128.  For linear probing, we freeze the encoder and train only a linear classifier with AdamW and a learning rate of $1\times10^{-4}$ for 100 epochs with a batch size of 128, and images resized to $224\times224$. Standard MAE augmentations, \eg, random resized crop and horizontal flip, are applied, and model selection is based on validation accuracy. Final results are reported on the test set using top-1 accuracy. \\

\noindent \textbf{Semantic Segmentation.}
We use GrowliFlower~\cite{growliflower} and the labelled split of PhenoBench~\cite{pnb} for downstream evaluation. The labelled PhenoBench split is provided separately by the PhenoBench authors from the unlabelled patches used for continual pretraining in AgriField-40K. PhenoBench provides a separate labelled set with dense semantic annotations for crop, weed, and soil/background classes. It includes UAV images of sugar beet fields in Meckenheim, Germany, with two sugar beet varieties and six weed species from multiple growth stages. Following the official benchmark, we use the standardized split of 1,407/772 for training and validation. GrowliFlower provides RGB UAV orthophoto patches of cauliflower fields near Cologne, Germany, with dense pixel-wise annotations. We use the GrowliFlowerL subset and the official splits. To mitigate severe class imbalance, we retain the background class and the three most frequent plant identifiers while discarding rare instance labels. For both datasets, we attach an UPerNet~\cite{upernet} head and fine-tune the adapters and head, keeping the backbone frozen. Training uses AdamW with learning rates of $1\times10^{-4}$ for adapters and $2\times10^{-4}$ for the decoder, weight decay of $1\times10^{-2}$, 80 epochs, and a batch size of 32, with all images resized to $512\times512$. Performance is evaluated on the PhenoBench validation set and GrowliFlower test set using class IoU and mean IoU. \\

\noindent \textbf{Object Detection.}
We evaluate on CropAndWeed~\cite{cropandweed}, using the CropOrWeed2 subset. It consists of RGB images collected from cultivation sites and experimental plots in Austria, with annotations for 74 crop and weed species. After removing images without valid annotations, the dataset contains 7,705 images, which we randomly split into 60\%/20\%/20\% training, validation, and test sets. We attach a Faster R-CNN~\cite{faster_rcnn} detector with a Feature Pyramid Network (FPN)~\cite{fpn} and fine-tune the adapters and detection head keeping the backbone frozen. We use AdamW with learning rates of $1\times10^{-5}$ for adapters and $1\times10^{-4}$ for the detection head, cosine learning rate decay, 50 epochs, and a batch size of 32. Images are resized to a shorter side of 800 pixels. Performance is evaluated on the test set using mAP@50 and mAP@50:95.

\subsection{Main Results}
\label{sec:results}

\noindent \textbf{Classification on DeepWeeds.}
Table~\ref{tab:deepweeds_test} reports DeepWeeds results under linear probing and PEFT adaptation. In the linear probing setting, AgriMAE improves over the original MAE, with feature reconstruction providing the strongest gains among the AgriMAE variants. A similar trend is observed under PEFT, where randomly initialized adapters already provide a strong baseline, but AgriMAE further improves downstream performance through continual pretraining. AgriMAE trained with $\mathcal{L}_{feat}$ performs best among the parameter-efficient methods and also surpasses full backbone fine-tuning. Importantly, this is achieved while updating almost $9\times$ fewer trainable parameters which demonstrates the effectiveness of parameter-efficient adaptation for limited-data settings. \\

\begin{table}[t]
    \centering
    \caption{DeepWeeds test-set classification performance. Linear probing trains only a linear head (middle), while PEFT fine-tunes adapters and a task-specific head and reports the number of trainable parameters (right). Results are averaged over five seeds. NA denotes non-applicable settings. We highlight the best result among parameter-efficient methods and include full fine-tuning as a strong reference point.}
    \label{tab:deepweeds_test}

    \resizebox{\textwidth}{!}{%
    \begin{tabular}{l|cc|cc|cc}
    \hline
    \multicolumn{1}{c|}{}
    & \multicolumn{2}{c|}{}
    & \multicolumn{2}{c|}{\textbf{Linear Probing}}
    & \multicolumn{2}{c}{\textbf{PEFT}} \\
    \cline{4-7}
    \textbf{Method}
    & $\mathcal{L}_{\mathrm{pix}}$
    & $\mathcal{L}_{\mathrm{feat}}$
    & Top-1 Acc. (\%)
    & Params. (M)
    & Top-1 Acc. (\%)
    & Params. (M) \\
    \hline

    \rowcolor{gray!5}
    \textcolor{gray}{Full Fine-tuning}
    & \textcolor{gray}{\checkmark}
    & \textcolor{gray}{--}
    & \textcolor{gray}{NA}
    & \textcolor{gray}{NA}
    & \textcolor{gray}{92.23$\pm$0.35}
    & \textcolor{gray}{85.81} \\

    \hline

    \rowcolor{gray!15}
    MAE (Frozen)
    & \checkmark
    & --
    & 86.04$\pm$0.10
    & 0.01
    & NA
    & NA \\

    \rowcolor{gray!15}
    MAE + Adapters
    & \checkmark
    & --
    & NA
    & NA
    & 93.66$\pm$0.12
    & 9.46 \\

    \rowcolor{gray!15}
    AgriMAE + Adapters
    & \checkmark
    & --
    & 86.84$\pm$0.22
    & 0.01
    & 94.15$\pm$0.07
    & 9.46 \\

    \rowcolor{gray!15}
    AgriMAE + Adapters
    & \checkmark
    & \checkmark
    & 88.25$\pm$0.15
    & 0.01
    & 94.47$\pm$0.14
    & 9.46 \\

    \rowcolor{gray!15}
    AgriMAE + Adapters
    & --
    & \checkmark
    & \textbf{88.63$\pm$0.15}
    & \textbf{0.01}
    & \textbf{94.72$\pm$0.20}
    & \textbf{9.46} \\
    \hline
    \end{tabular}
    }
\end{table}

\begin{figure}[t]
	\centering

    \subfloat{\includegraphics[scale=0.21]{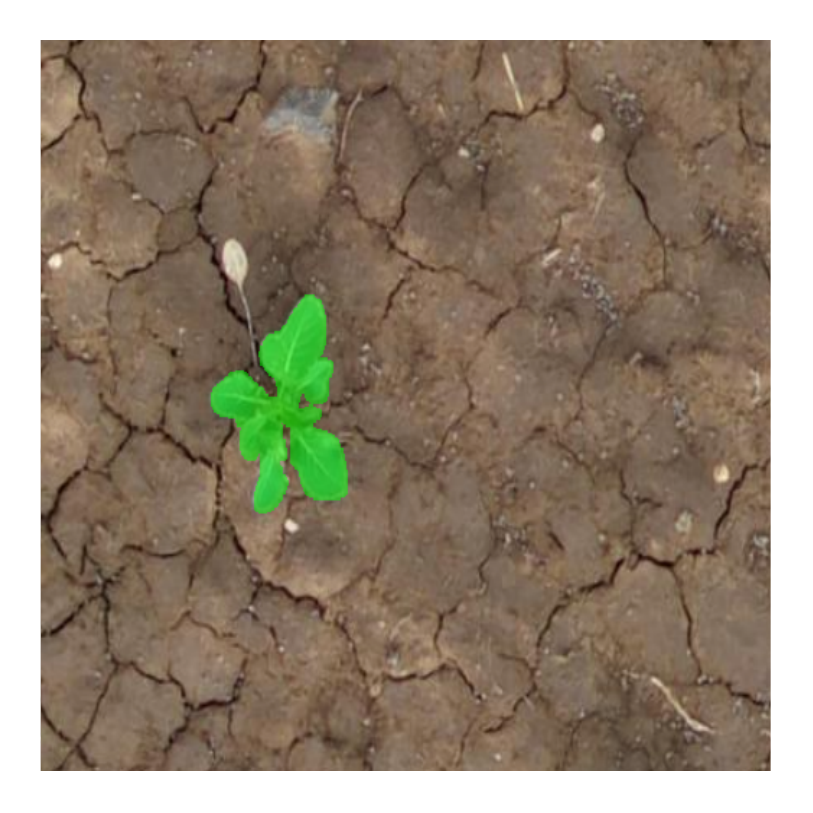}}\hfill
	\subfloat{\includegraphics[scale=0.21]{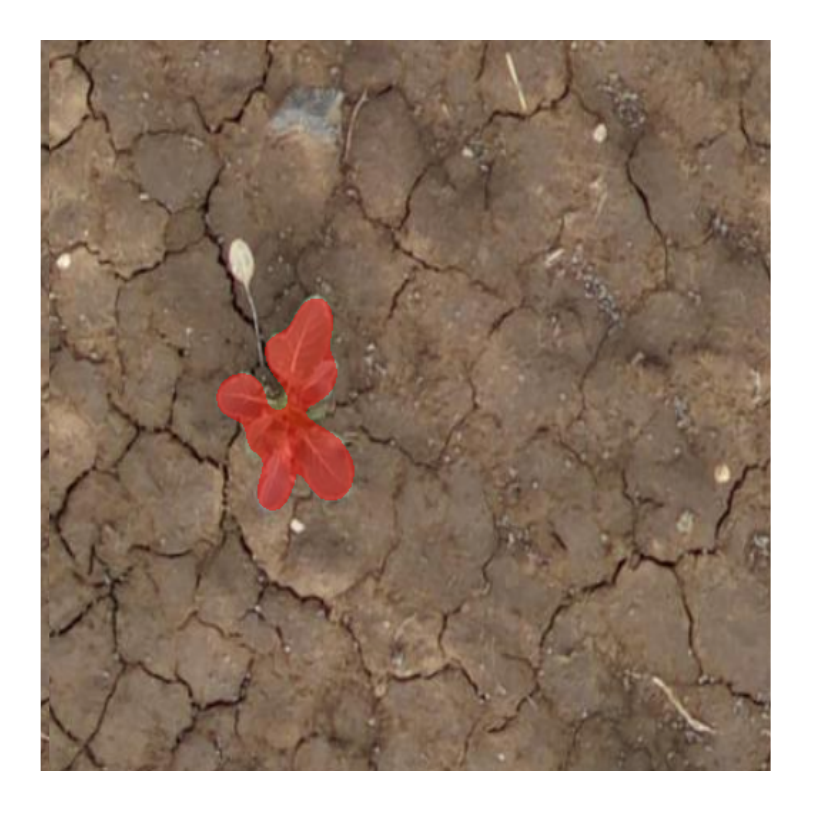}}\hfill 
    \subfloat{\includegraphics[scale=0.21]{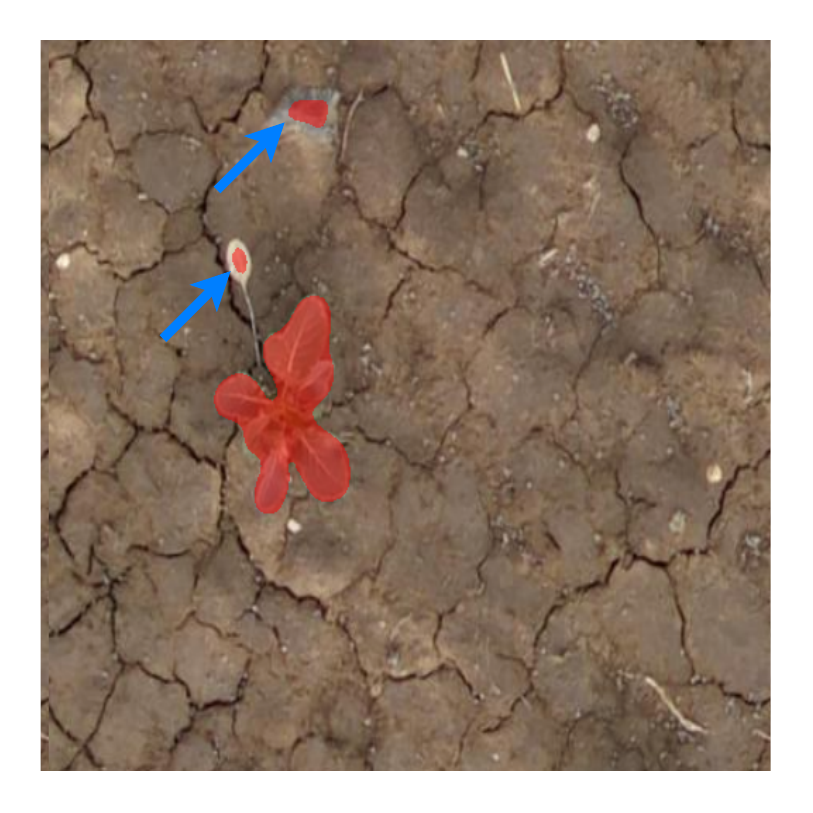}}\hfill 
    \subfloat{\includegraphics[scale=0.21]{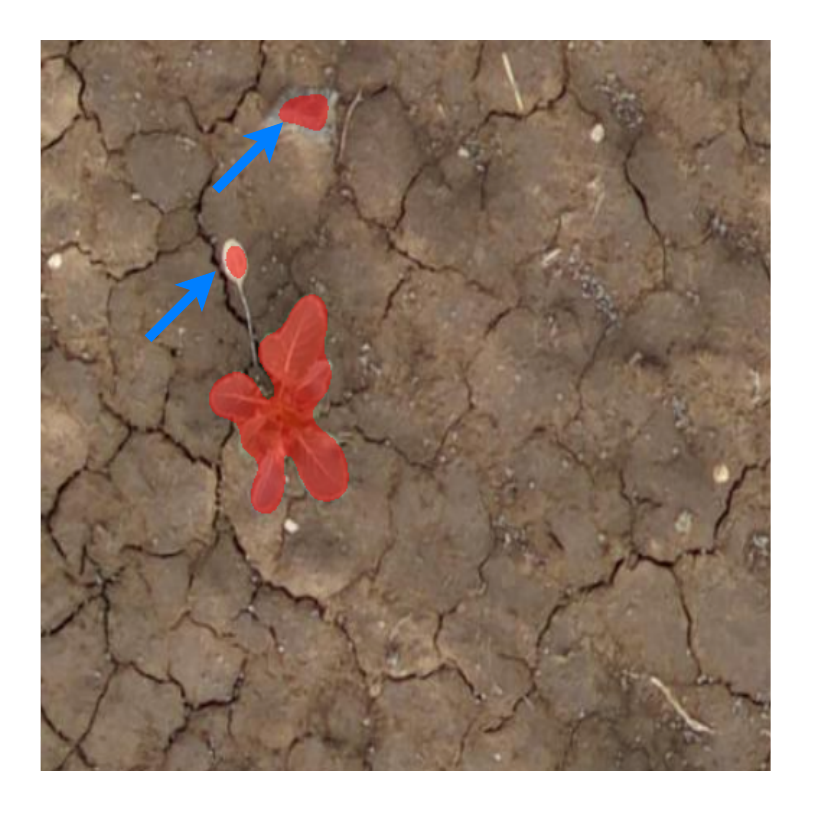}}\hfill

    \vspace{-1.0em}
    
	\subfloat{\includegraphics[scale=0.21]{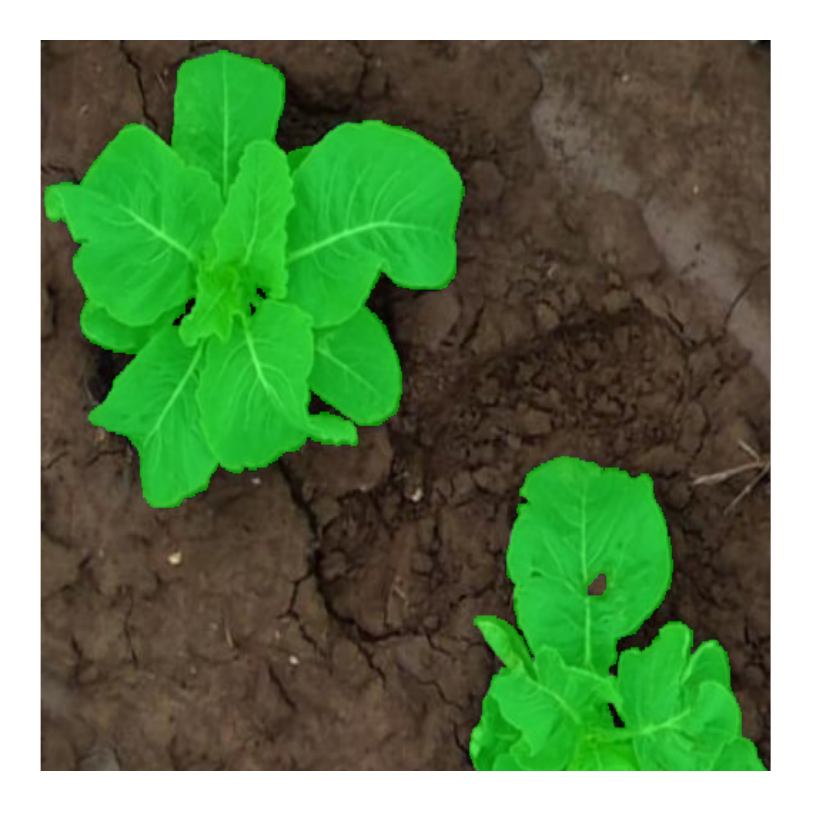}}\hfill 
	\subfloat{\includegraphics[scale=0.21]{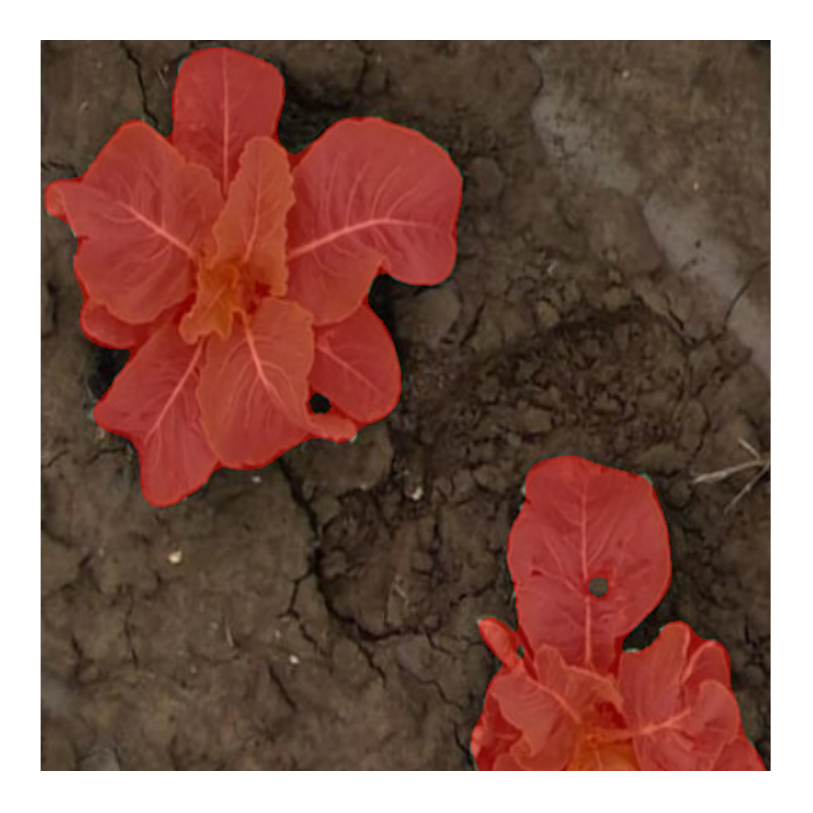}}\hfill 
    \subfloat{\includegraphics[scale=0.21]{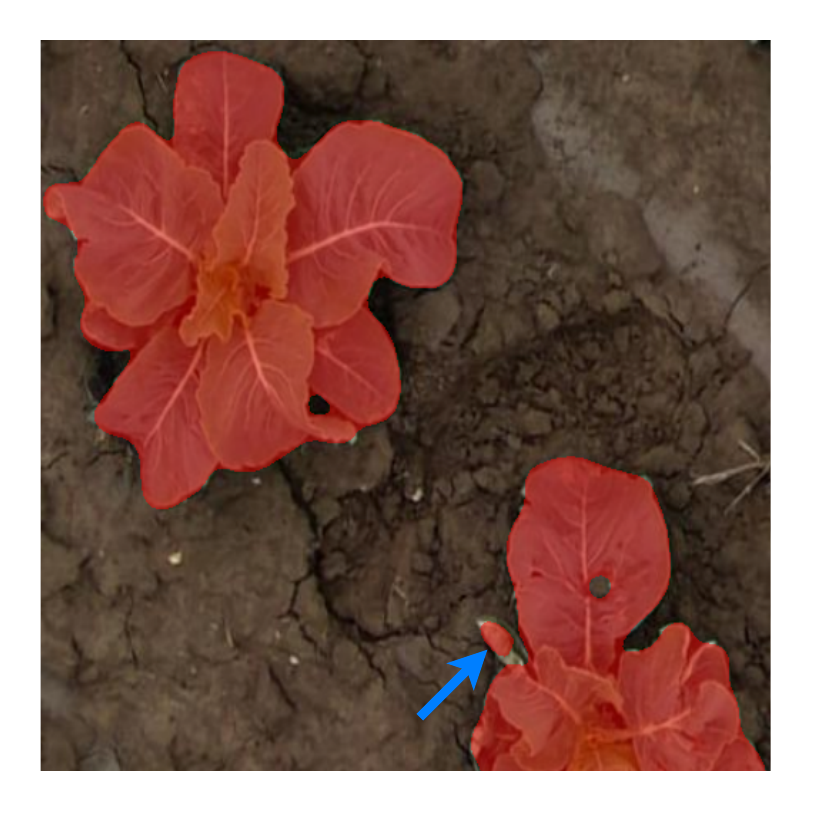}}\hfill 
    \subfloat{\includegraphics[scale=0.21]{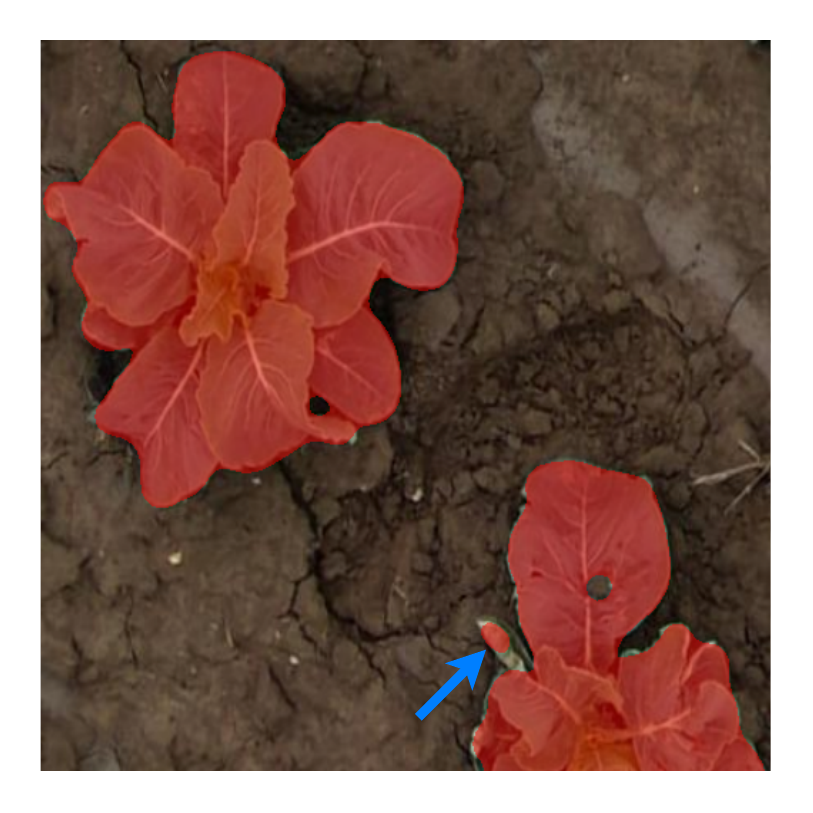}}\hfill

    \vspace{0em}
    \noindent\rule{\linewidth}{0.01pt}
    \vspace{0em}

	\subfloat{\includegraphics[scale=0.21]{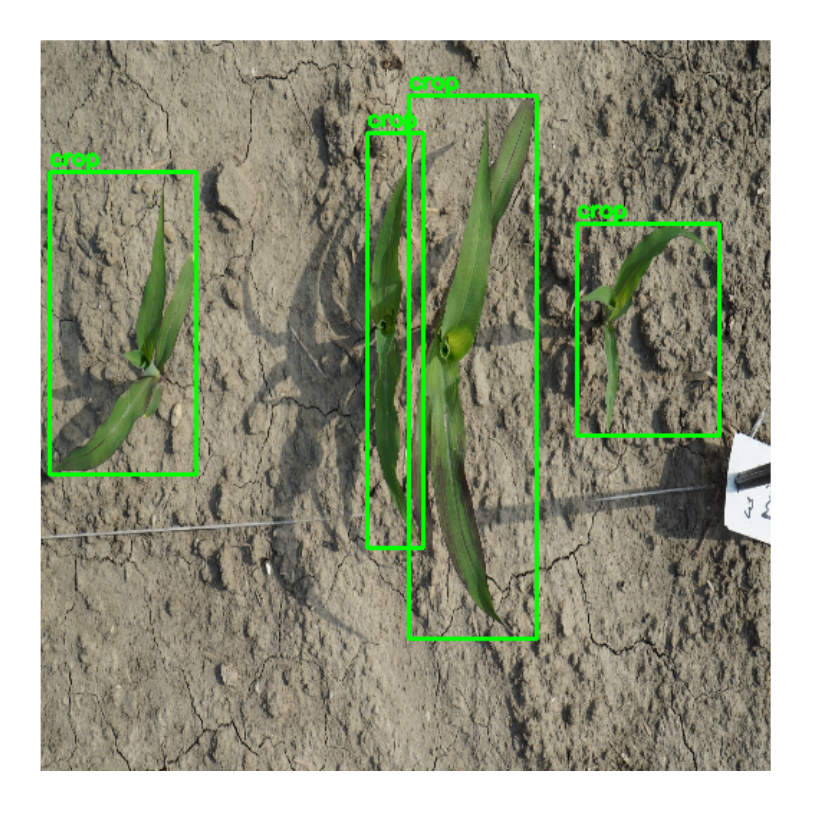}}\hfill
	\subfloat{\includegraphics[scale=0.21]{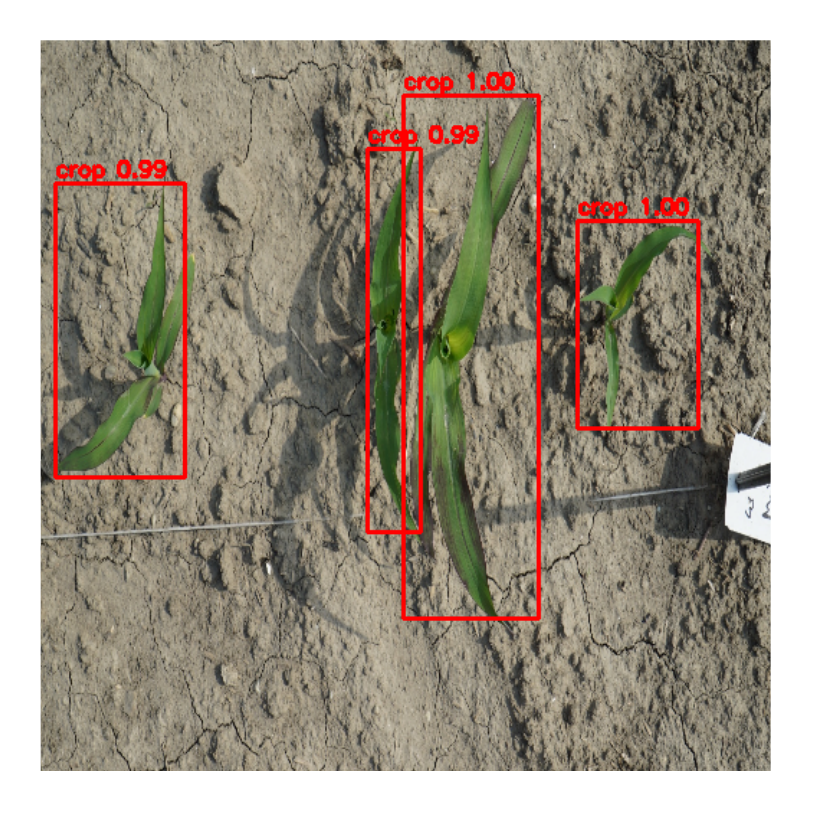}}\hfill 
    \subfloat{\includegraphics[scale=0.21]{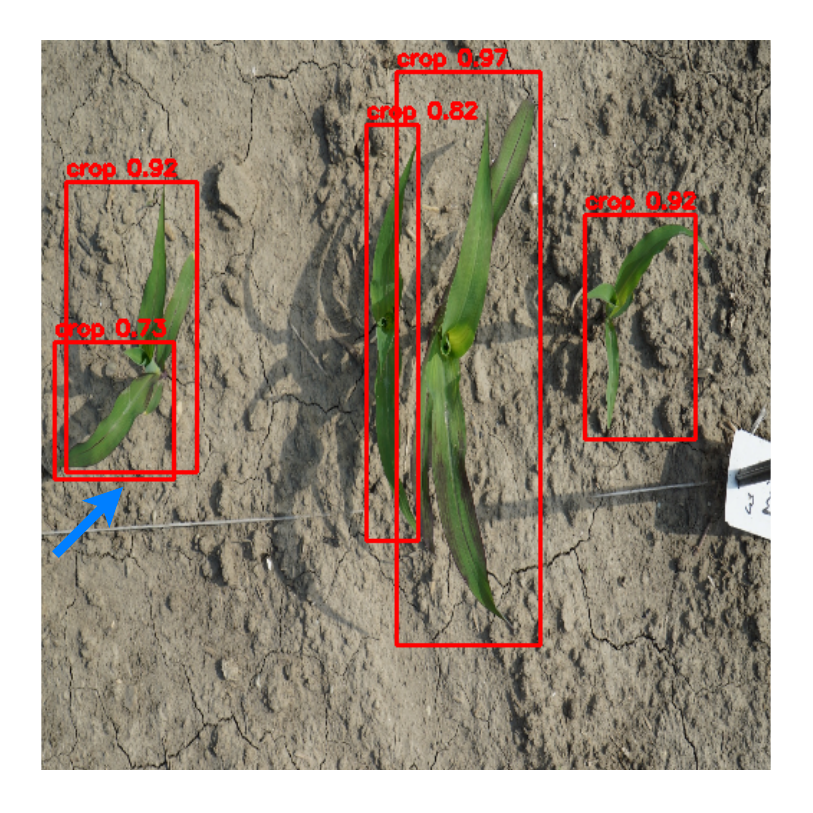}}\hfill 
    \subfloat{\includegraphics[scale=0.21]{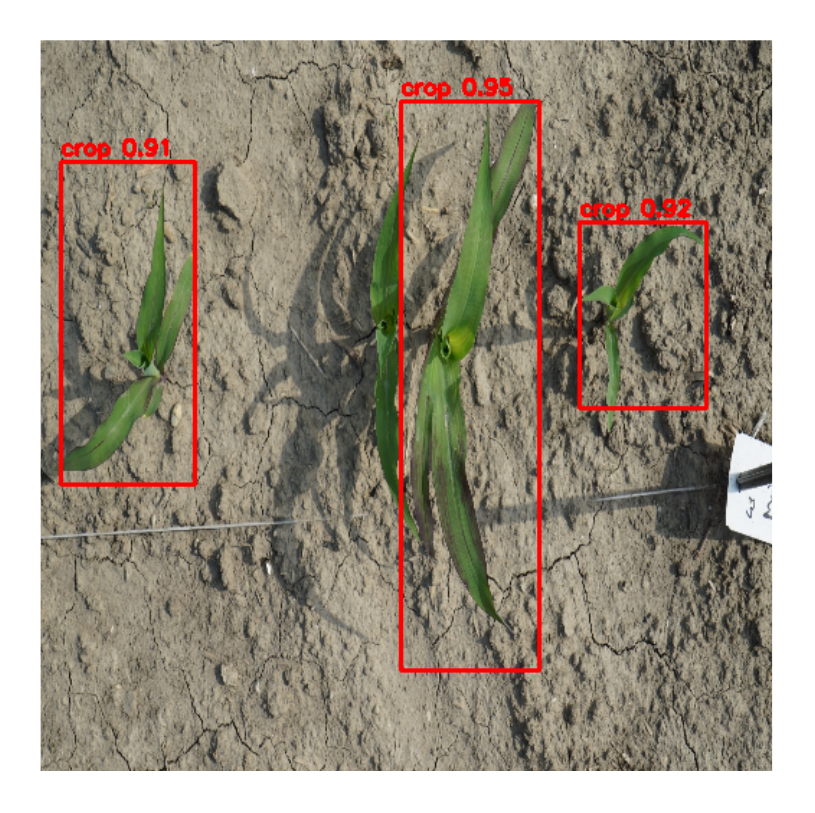}}\hfill 

    \vspace{-1.0em}

    \setcounter{subfigure}{0}
	\subfloat[\tiny Ground truth]{\includegraphics[scale=0.21]{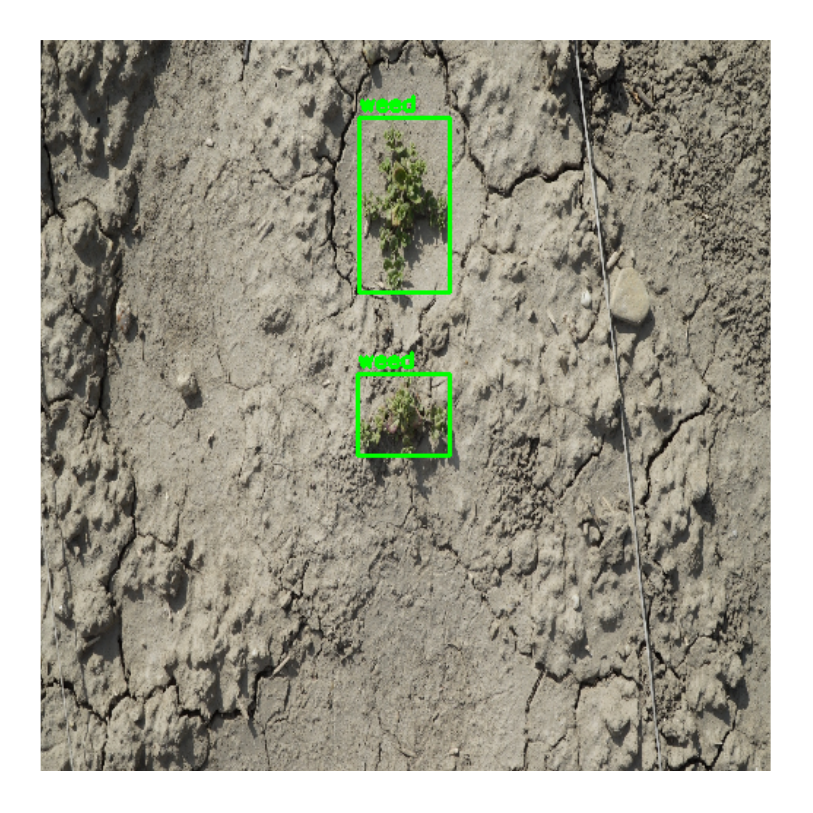}}\hfill 
	\subfloat[\tiny AgriMAE ($\mathcal{L}_{feat}$)]{\includegraphics[scale=0.21]{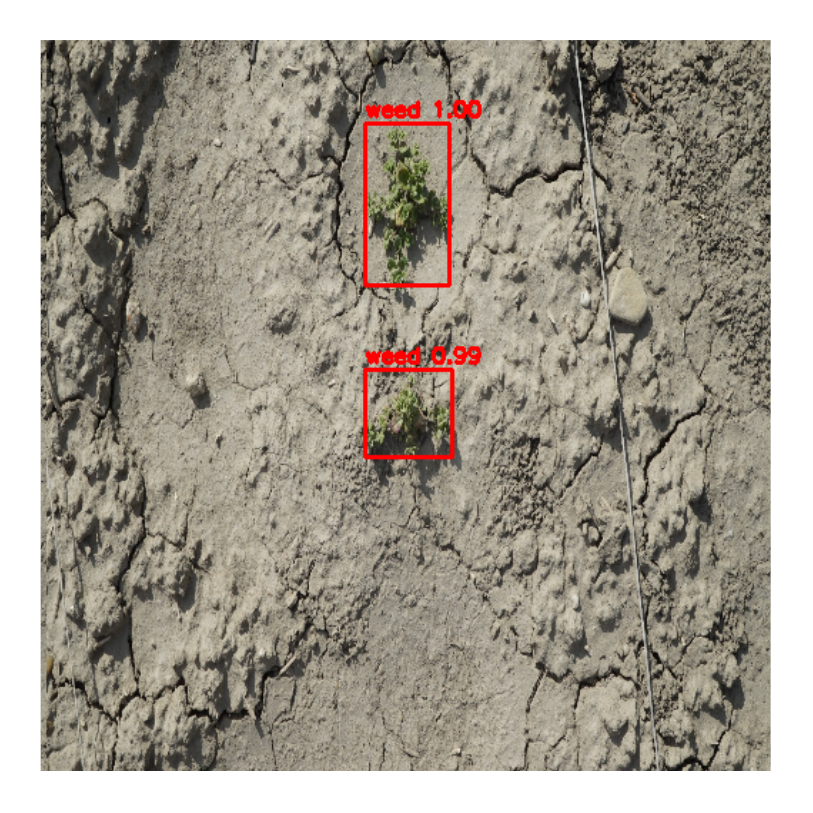}}\hfill 
    \subfloat[\tiny AgriMAE ($\mathcal{L}_{pix}$)]{\includegraphics[scale=0.21]{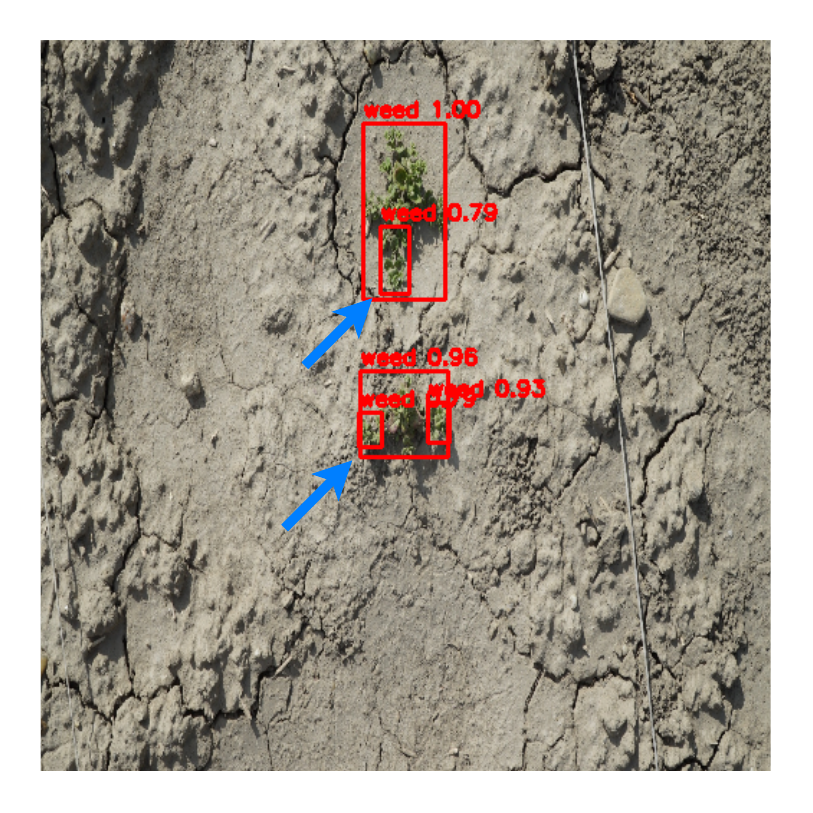}}\hfill 
    \subfloat[\tiny MAE]{\includegraphics[scale=0.21]{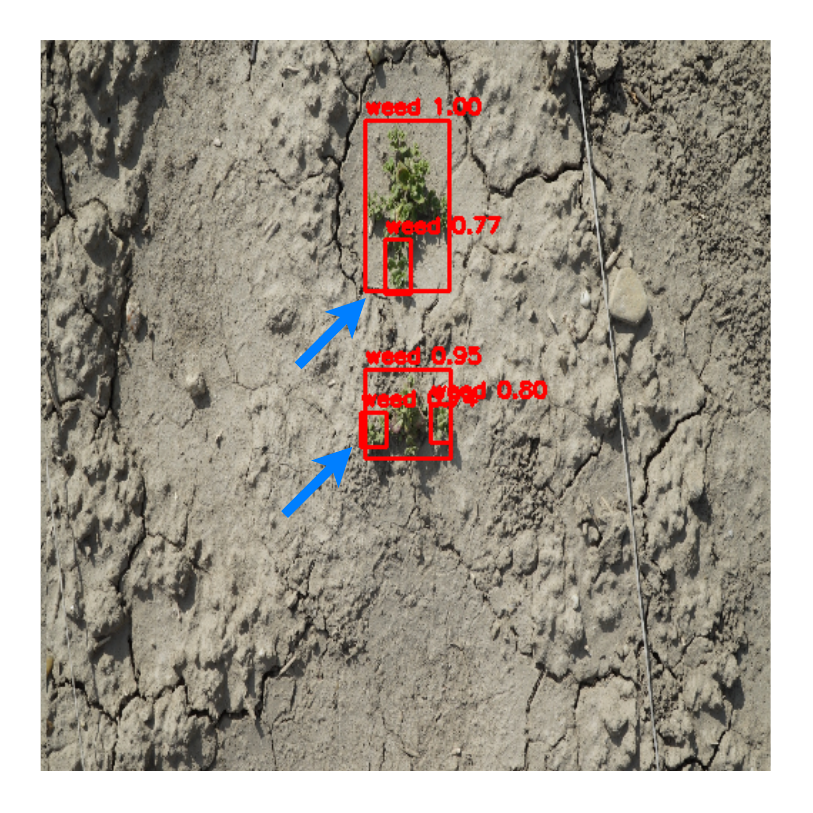}}\hfill 

    \caption{Qualitative results. Top: Examples for the semantic segmentation task on GrowliFlower. Bottom: Examples for the object detection task on CropAndWeed.}
     \label{fig:examples}
\end{figure}

\noindent \textbf{Semantic Segmentation on PhenoBench.} Table~\ref{tab:phenobench_results} reports per-class IoU and mIoU on the validation set. Randomly initialized adapters achieve strong soil performance but lower performance on the more challenging weed class.  
Continual pretraining improves performance, and AgriMAE with $\mathcal{L}_{feat}$ achieves the best IoU on weed class. Full fine-tuning obtains slightly better performance on average but updates 121.06M parameters compared to 44.08M for adapters. AgriMAE therefore almost matches full fine-tuning with fewer trainable parameters while improving the most challenging weed class. \\

\begin{table}[t]
    \centering
    \caption{Segmentation results on the PhenoBench~\cite{pnb} validation set. All models use an UPerNet~\cite{upernet} head. Results are averaged over five seeds. We highlight the best result among efficient methods and include full fine-tuning as a strong reference point.}
    \label{tab:phenobench_results}

    \resizebox{\textwidth}{!}{%
    \begin{tabular}{l|cc|ccc|c|c}
    \hline
    \multicolumn{1}{c|}{}
    & \multicolumn{2}{c|}{}
    & \multicolumn{1}{c}{\textbf{Soil}}
    & \multicolumn{1}{c}{\textbf{Crop}}
    & \multicolumn{1}{c|}{\textbf{Weed}}
    & \textbf{Overall}
    & \textbf{Model} \\
    \cline{4-8}
    \textbf{Method}
    & $\mathcal{L}_{\mathrm{pix}}$
    & $\mathcal{L}_{\mathrm{feat}}$
    & 
    & \textbf{IoU (\%)}
    & 
    & \textbf{mIoU (\%)}
    & \textbf{Params. (M)} \\
    \hline

    \rowcolor{gray!5}
    \textcolor{gray}{Full Fine-tuning}
    & \textcolor{gray}{\checkmark}
    & \textcolor{gray}{--}
    & \textcolor{gray}{98.50$\pm$0.02}
    & \textcolor{gray}{86.75$\pm$0.65}
    & \textcolor{gray}{57.14$\pm$0.36}
    & \textcolor{gray}{80.80$\pm$0.11}
    & \textcolor{gray}{121.06} \\

    \hline

    \rowcolor{gray!15}
    MAE + Adapters
    & \checkmark
    & --
    & 98.33$\pm$0.05
    & 85.20$\pm$0.18
    & 55.90$\pm$0.31
    & 79.81$\pm$0.10
    & 44.08 \\

    \rowcolor{gray!15}
    AgriMAE + Adapters
    & \checkmark
    & --
    & 98.33$\pm$0.04
    & 85.67$\pm$0.13
    & 56.22$\pm$0.33
    & 80.07$\pm$0.07
    & 44.08 \\

    \rowcolor{gray!15}
    AgriMAE + Adapters
    & --
    & \checkmark
    & \textbf{98.35$\pm$0.05}
    & \textbf{86.12$\pm$0.11}
    & \textbf{57.28$\pm$0.24}
    & \textbf{80.58$\pm$0.10}
    & \textbf{44.08} \\
    \hline
    \end{tabular}
    }
\end{table}

\noindent \textbf{Semantic Segmentation on GrowliFlower.} Table~\ref{tab:growliflower_results} shows per-class IoU and mIoU on the test set. AgriMAE consistently improves over the original MAE baseline, with feature reconstruction producing better fine-grained segmentation details, as shown in the upper part of Fig.~\ref{fig:examples}. 
We observe higher variance for under-represented classes, showing the strong class imbalance in the dataset. Class 3 remains very challenging even with full fine-tuning, likely due to limited representation (530 images compared to 1,381 and 1,211 for classes 1 and 2). \\

\begin{table}[t]
    \centering
    \caption{Semantic segmentation results on the GrowliFlower~\cite{growliflower} test set. All models use an UPerNet~\cite{upernet} head. Results are averaged over five seeds. We highlight the best result among efficient methods and include full fine-tuning as a strong reference point.}
    \label{tab:growliflower_results}

    \resizebox{\textwidth}{!}{%
    \begin{tabular}{l|cc|cccc|c|c}
    \hline
    \multicolumn{1}{c|}{}
    & \multicolumn{2}{c|}{}
    & \textbf{0}
    & \textbf{1}
    & \textbf{2}
    & \textbf{3}
    & \textbf{Overall}
    & \textbf{Model} \\
    \cline{4-9}
    \textbf{Method}
    & $\mathcal{L}_{\mathrm{pix}}$
    & $\mathcal{L}_{\mathrm{feat}}$
    & \multicolumn{4}{c|}{\textbf{IoU (\%)}}
    & \textbf{mIoU (\%)}
    & \textbf{Params. (M)} \\
    \hline

    \rowcolor{gray!5}
    \textcolor{gray}{Full Fine-tuning}
    & \textcolor{gray}{\checkmark}
    & \textcolor{gray}{--}
    & \textcolor{gray}{98.97$\pm$0.03}
    & \textcolor{gray}{68.68$\pm$0.41}
    & \textcolor{gray}{51.35$\pm$1.24}
    & \textcolor{gray}{20.93$\pm$0.99}
    & \textcolor{gray}{59.93$\pm$0.46}
    & \textcolor{gray}{121.06} \\

    \hline

    \rowcolor{gray!15}
    MAE + Adapters
    & \checkmark
    & --
    & 98.83$\pm$0.12
    & 66.28$\pm$0.65
    & 44.74$\pm$1.26
    & 16.84$\pm$0.37
    & 56.58$\pm$0.78
    & 44.08 \\

    \rowcolor{gray!15}
    AgriMAE + Adapters
    & \checkmark
    & --
    & 98.94$\pm$0.01
    & 67.26$\pm$0.58
    & 46.64$\pm$1.54
    & 16.04$\pm$0.80
    & 57.22$\pm$0.46
    & 44.08 \\

    \rowcolor{gray!15}
    AgriMAE + Adapters
    & --
    & \checkmark
    & \textbf{98.97$\pm$0.03}
    & \textbf{67.89$\pm$0.44}
    & \textbf{48.12$\pm$0.77}
    & \textbf{17.39$\pm$1.03}
    & \textbf{58.09$\pm$0.35}
    & \textbf{44.08} \\
    \hline

    \end{tabular}
    }
\end{table}

\noindent \textbf{Object Detection on CropAndWeed.} Table~\ref{tab:cropandweed_results} reports object detection results on the CropAndWeed test set. MAE provides a solid baseline, and continual pretraining improves both mAP@50 and mAP@50:95. AgriMAE with $\mathcal{L}_{feat}$ provides the strongest performance, significantly outperforming full fine-tuning across all detection metrics. This is achieved with about 74\% fewer trainable parameters, showing the effectiveness of semantic reconstruction for parameter-efficient settings. As shown in the lower part of Figure~\ref{fig:examples}, AgriMAE with $\mathcal{L}_{feat}$ detects all instances correctly, whereas  MAE produces several false positives.

\begin{table}[t]
    \centering
    \caption{Object detection results on the CropAndWeed~\cite{cropandweed} test set. All models use a Faster R-CNN~\cite{faster_rcnn} head. Results are averaged over five seeds. We highlight the best result among efficient methods and include full fine-tuning as a strong reference point.}
    \label{tab:cropandweed_results}

    \resizebox{\textwidth}{!}{%
    \begin{tabular}{l|cc|cccc|cc|c}
    \hline
    \multicolumn{1}{c|}{}
    & \multicolumn{2}{c|}{}
    & \multicolumn{2}{c|}{\textbf{Crop}}
    & \multicolumn{2}{c|}{\textbf{Weed}}
    & \multicolumn{2}{c|}{\textbf{Overall}}
    & \textbf{Model} \\
    \cline{4-10}
    \textbf{Method}
    & $\mathcal{L}_{\mathrm{pix}}$
    & $\mathcal{L}_{\mathrm{feat}}$
    & \textbf{AP$_{50}$}
    & \textbf{AP$_{50:95}$}
    & \textbf{AP$_{50}$}
    & \textbf{AP$_{50:95}$}
    & \textbf{mAP$_{50}$}
    & \textbf{mAP$_{50:95}$}
    & \textbf{Params. (M)} \\
    \hline
    \rowcolor{gray!5}
    \textcolor{gray}{Full Fine-tuning}
    & \textcolor{gray}{\checkmark}
    & \textcolor{gray}{--}
    & \textcolor{gray}{77.26$\pm$0.60}
    & \textcolor{gray}{48.67$\pm$0.76}
    & \textcolor{gray}{49.17$\pm$0.43}
    & \textcolor{gray}{24.36$\pm$0.28}
    & \textcolor{gray}{63.25$\pm$0.31}
    & \textcolor{gray}{36.44$\pm$0.26}
    & \textcolor{gray}{103.46} \\

    \hline
    \rowcolor{gray!15}
    MAE + Adapters
    & \checkmark
    & --
    & 72.54$\pm$0.65
    & 45.33$\pm$0.63
    & 47.51$\pm$0.32
    & 23.27$\pm$0.41
    & 60.03$\pm$0.48
    & 34.30$\pm$0.47
    & 27.11 \\

    \rowcolor{gray!15}
    AgriMAE + Adapters
    & \checkmark
    & --
    & 73.37$\pm$0.36
    & 46.56$\pm$0.31
    & 48.31$\pm$0.52
    & 23.79$\pm$0.52
    & 60.79$\pm$0.38
    & 35.18$\pm$0.36
    & 27.11 \\

    \rowcolor{gray!15}
    AgriMAE + Adapters
    & --
    & \checkmark
    & \textbf{79.92$\pm$0.82}
    & \textbf{52.02$\pm$0.50}
    & \textbf{52.54$\pm$0.88}
    & \textbf{27.35$\pm$0.58}
    & \textbf{66.23$\pm$0.72}
    & \textbf{39.68$\pm$0.51}
    & \textbf{27.11} \\
    \hline

    \end{tabular}
    }
\end{table}

\subsection{Ablation Studies}
\label{sec:ablation}

\noindent \textbf{Adapter Bottleneck Rank.} We study the effect of the AdaptFormer~\cite{adaptformer} bottleneck rank $r$ using linear probing on the DeepWeeds validation set. We evaluate $r \in \{32, 64, 128, 256, 512\}$ while keeping all other continual pretraining settings fixed. As shown in Tab.~\ref{tab:ablation}, linear probing performance improves consistently with increasing $r$. We therefore use $r=512$ as a default in all of our experiments. Although $r=512$ is relatively large compared to the ViT-B hidden dimension ($d=768$), adapters remain substantially more efficient than full fine-tuning. \\

\noindent \textbf{Reconstruction Objective.}
We ablate the effect of pixel and feature reconstruction in continual pretraining, and their combination using linear probing on the DeepWeeds validation set. As shown in Tab.~\ref{tab:ablation}, AgriMAE with $\mathcal{L}_{pix}$ improves over the frozen backbone, demonstrating the benefit of continual pretraining on AgriField-40K. Combining this with $\mathcal{L}_{feat}$ further improves accuracy, while AgriMAE trained only with $\mathcal{L}_{feat}$ achieves the best results. 

\begin{table}[t]
    \centering
    \caption{Ablation study for linear probing on the DeepWeeds validation set. Left: Effect of the adapter bottleneck dimension $r$ for AgriMAE trained with $\mathcal{L}_{\mathrm{pix}}$, shown from a single run. Right: Effect of different pretraining objectives, evaluated with the best $r=512$ and reported as the mean and standard deviation over five seeds.}
    \label{tab:ablation}

    \resizebox{0.85\textwidth}{!}{%
    \begin{tabular}{c | c @{\hspace{1.5em}} || l | cc| c}
    \hline
    \multicolumn{2}{c}{\textbf{Bottleneck} ($\mathcal{L}_{\mathrm{pix}}$)}
    &
    \multicolumn{4}{c}{\textbf{Pretraining Objective} ($r=512$)} \\
    \hline
    \textbf{$r$}
    & Top-1 Acc. (\%)
    &
    Method
    & $\mathcal{L}_{\mathrm{pix}}$
    & $\mathcal{L}_{\mathrm{feat}}$
    & Top-1 Acc. (\%) \\
    \hline
    32
    & 83.57
    &
    &
    &
    &
    \\

    64
    & 84.85
    &
    MAE (Frozen)
    & \checkmark
    & --
    & 85.34$\pm$0.13 \\

    128
    & 85.78
    &
    AgriMAE + Adapters
    & \checkmark
    & --
    & 86.52$\pm$0.10 \\

    256
    & 86.09
    &
    AgriMAE + Adapters
    & \checkmark
    & \checkmark
    & 87.58$\pm$0.12 \\

    512
    & \textbf{86.52}
    &
    AgriMAE + Adapters
    & --
    & \checkmark
    & \textbf{87.82$\pm$0.27} \\
    \hline
    \end{tabular}
    }
\end{table}

\section{Discussion}
\label{sec:Conclusion}

Our results demonstrate the effectiveness of parameter-efficient continual pretraining for field-centric agricultural vision. Across classification, semantic segmentation, and object detection, AgriMAE improves over the original MAE adapter baseline, showing that adapting representations on AgriField-40K benefits downstream tasks. A key finding is that the reconstruction target matters. While pixel reconstruction already improves over the ImageNet-pretrained MAE, semantic feature reconstruction provides larger gains. As shown in Fig.~\ref{fig:tsne_comparison}, our feature-reconstruction variant produces more compact and better separated clusters. This suggests that leveraging a strong feature extractor such as DINOv3 for dense feature-level guidance is useful for field imagery, where crop, weed, and background structure is often distributed across the scene. The qualitative examples in Fig.~\ref{fig:examples} show the same trend, with feature reconstruction producing cleaner segmentation masks and more accurate detections than pixel reconstruction. Finally, our ablation shows that larger adapter ranks strengthen continual pretraining, indicating that AgriMAE benefits from added adaptation capacity while still remaining far more parameter-efficient than full fine-tuning. \\

To keep the study controlled, we focus on masked image modelling and compare PEFT adaptation of the ImageNet-pretrained MAE~\cite{mae} with AgriMAE on several downstream tasks. While this isolates the effect of continual pretraining, future work could explore other representation learning approaches, such as contrastive learning~\cite{simclr}, self-distillation~\cite{dino}, and diffusion-based representation learning~\cite{xiang2026sprout}. Moreover, we implement AgriMAE with a ViT-B~\cite{vit} backbone, AdaptFormer~\cite{adaptformer}, and DINOv3~\cite{dinov3} ViT-L feature targets to provide a controlled and reproducible baseline. Exploring alternative teacher feature extractors for semantic reconstruction, or extending our adaptation framework to larger backbones, and evaluating other PEFT variants such as LoRA~\cite{lora}, and ExPLoRA~\cite{explora} are promising directions. Future work could also study efficient adaption of agriculture-specific foundation models~\cite{agrifm,shen2025weednet,xiang2026sprout}, with AgriField-40K serving as a field-centric resource for efficient continual pretraining. Finally, three of our downstream evaluation tasks are fully independent of the AgriField-40K pretraining corpus. For PhenoBench, we included its unlabelled patches in AgriField-40K for continual pretraining and used its officially labelled split for downstream supervised fine-tuning, though PhenoBench does not provide sufficient detail on how images were partitioned across fields, sequences, or campaigns. This follows common evaluation practice in parameter-efficient continual pretraining~\cite{explora}, where unlabelled splits are used for pretraining and labelled splits from the same dataset are used for downstream evaluation. \\

\noindent \textbf{Conclusion.}
We introduce AgriField-40K and establish AgriMAE as a strong parameter-efficient baseline for continual pretraining. Our results show that AgriMAE can match or even outperform full fine-tuning while using substantially fewer trainable parameters. We hope AgriField-40K will support future research on efficient continual pretraining and practical agricultural vision applications.


\section*{Acknowledgements}
This work was supported by Innovation Fund Denmark under Grant Agreement No. 2105-00013A (SAVA - Safety in Autonomous Vehicles in Agriculture) and the Novo Nordisk Foundation under Grant Agreement No. NNF25SA0104538 (RIC - Robotic Intercropping). HPC resources for the initial experiments were provided by the Pioneer Centre for Artificial Intelligence and the DTU Computing Center~\cite{DTU_DCC_resource}.

\bibliographystyle{splncs04}
\bibliography{main}
\end{document}